\documentclass[10pt,twocolumn,letterpaper]{article}
\usepackage[dvipsnames,table]{xcolor}
\usepackage{tablefootnote}

\usepackage[pagenumbers]{iccv} 

\usepackage{xcolor}

\definecolor{iccvblue}{rgb}{0.21,0.49,0.74}
\usepackage[pagebackref,breaklinks,colorlinks,allcolors=iccvblue]{hyperref}

\def\paperID{9662} 
\def\confName{ICCV}
\def\confYear{2025}

\title{LLaVA-SP: Enhancing Visual Representation with Visual Spatial Tokens for
MLLMs}

\author{
Haoran Lou\textsuperscript{1,}\textsuperscript{\textdagger} \quad
Chunxiao Fan\textsuperscript{1,}\textsuperscript{\textdagger} \quad
Ziyan Liu\textsuperscript{1} \quad
Yuexin Wu\textsuperscript{1} \quad
Xinliang Wang\textsuperscript{2} \\
\textsuperscript{1}Beijing University of Posts and Telecommunications  \quad
\textsuperscript{2}Beihang University \\
{\tt\small \{faker,cxfan,liuziyan,wuyuexin\}@bupt.edu.cn} \quad \tt\small wangxinliang@buaa.edu.cn
}

\begin{document}
\maketitle
\renewcommand{\thefootnote}{\textdagger}
\footnotetext{Corresponding author.}
\renewcommand{\thefootnote}{\arabic{footnote}}
\begin{abstract}

The architecture of multimodal large language models (MLLMs) commonly connects a vision encoder, often based on CLIP-ViT, to a large language model. While CLIP-ViT works well for capturing global image features, it struggles to model local relationships between adjacent patches, leading to weaker visual representation, which in turn affects the detailed understanding ability of MLLMs. To solve this, we propose LLaVA-SP, which \textbf{ only adds six spatial visual tokens} to the original visual tokens to enhance the visual representation. Our approach offers three key advantages: 1) We propose a novel Projector, which uses convolutional kernels to derive visual spatial tokens from ViT patch features, simulating two visual spatial ordering approaches: ``from central region to global" and ``from abstract to specific". Then, a cross-attention mechanism is applied to fuse fine-grained visual information, enriching the overall visual representation. 
2) We present two model variants:  LLaVA-SP-Cropping, which focuses on detail features through progressive cropping, and LLaVA-SP-Pooling, which captures global semantics through adaptive pooling, enabling the model to handle diverse visual understanding tasks. 3) Extensive experiments show that LLaVA-SP, fine-tuned with LoRA, achieves significant performance improvements across various multimodal benchmarks, outperforming the state-of-the-art LLaVA-1.5 model in multiple tasks with nearly identical inference latency. The code and models are available at \href{https://github.com/CnFaker/LLaVA-SP}{\texttt{https://github.com/CnFaker/LLaVA-SP}}.
\end{abstract}    
\section{Introduction}
\label{sec:intro}

\begin{figure}[t]
  \centering
\includegraphics[width=1.0\linewidth]{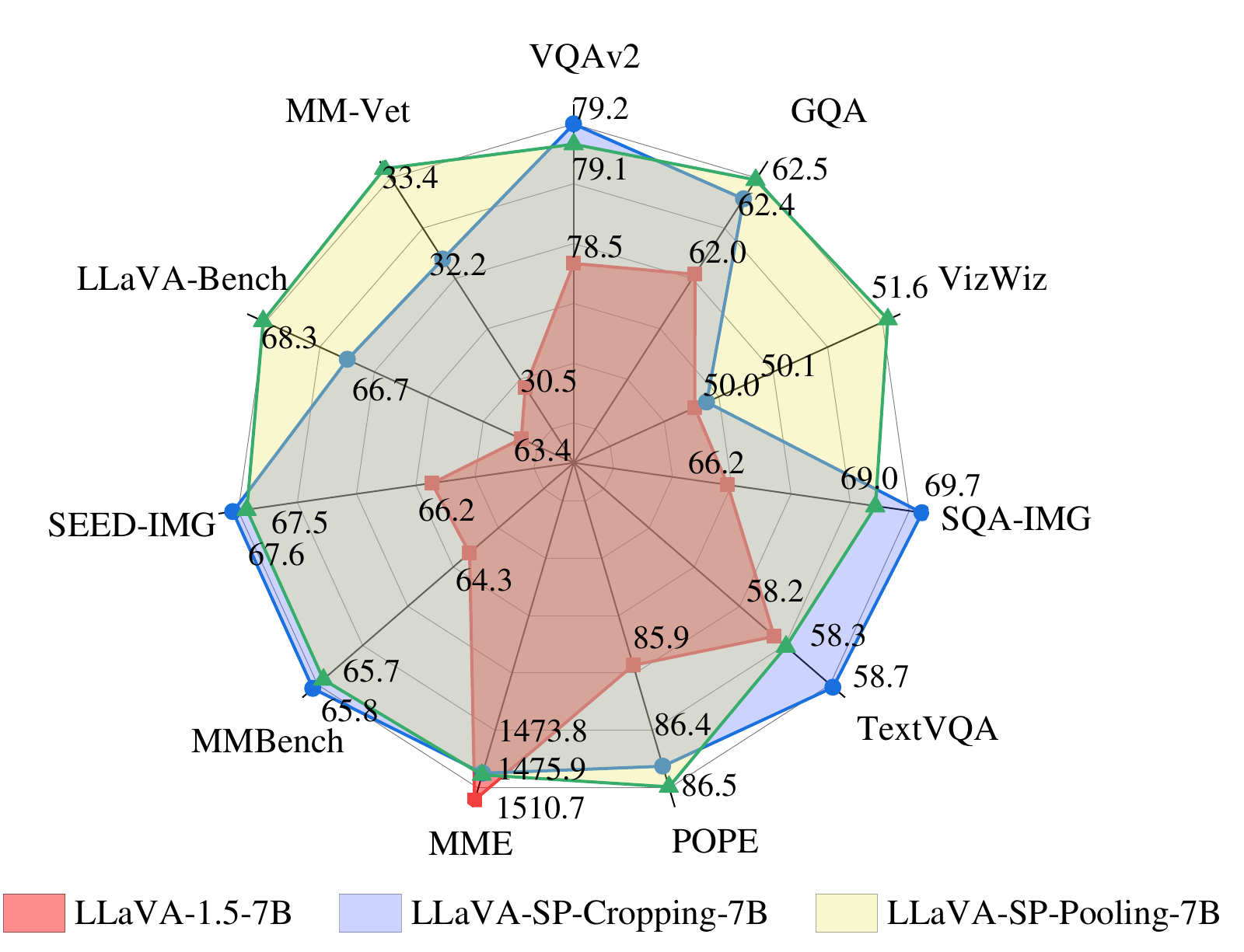}
   \caption{ Our models, \textbf{fine-tuned with LoRA}, outperform the \textbf{fully trained} LLaVA-1.5 in 10 out of 11 multimodal benchmarks. We name the model that employs the cropping operation as LLaVA-SP-Cropping and the one that uses the pooling operation as LLaVA-SP-Pooling.}
   \label{eval}
\end{figure}

 Multimodal large language models (MLLMs)~\cite{minigpt4,llava,llava1.5,internvl1.5,Qwen2VL,Qwen-VL} demonstrate exceptional capabilities in understanding visual and linguistic information, with the key to cross-modal understanding being modality alignment~\cite{yuan1,yuan2,yuan3,yuan4,liang1}. Recent research on aligning visual and language representation in MLLMs has primarily focused on the visual aspect. To reduce hallucinations in MLLMs caused by visual content, various strategies have been employed, such as increasing image resolution, using more powerful vision encoder, and integrating multiple visual features. For instance, LLaVA-1.5~\cite{llava1.5} increased input image resolution to 336, while InternVL-1.5~\cite{internvl1.5} proposed a dynamic high-resolution image strategy that supports 1024-resolution image inputs. SPHINX~\cite{sphinx} combined multiple vision encoders to extract diverse visual features. Monkey~\cite{monkey} fed different image blocks in parallel to their respective ViT encoders~\cite{vit} to learn unique features. Mini-Gemini~\cite{minigemini} proposed simultaneously inputting low-resolution and high-resolution images into the visual model. However, these approaches often lead to increased visual token counts, resulting in significantly increased training and inference costs.

Currently, mainstream MLLMs utilize CLIP-ViT~\cite{clip} as their vision encoder, but CLIP-ViT faces two limitations: 1) The contrastive learning paradigm relies on noisy image-text pair datasets during training, which limits its ability to understand fine-grained perceptual details. 2) ViT~\cite{vit} splits 2D images into flattened 1D patches, disrupting the intrinsic spatial relationships among adjacent patches. Research~\cite{vitcomer} indicates that while ViT is adept at capturing global information, it struggles to model the local relationships between neighboring patches. 

Based on the discussion above, this paper proposes a question: \emph{Can we fully leverage the capabilities of the vision encoder to enhance visual feature representation without significantly increasing the number of visual tokens}?

To address this question, we propose LLaVA-SP to enhance the visual representation of MLLMs. The Projector of LLaVA-SP consists of two key designs: the \textbf{Spatial Feature Extractor (SFE)} and the \textbf{Detail Feature Integrator (DFI)}.
1) The SFE aims to enhance the feature representation of the vision encoder by adding only six visual spatial tokens. These six visual spatial tokens can be introduced through two operations: cropping or pooling. The motivation for cropping is to emphasize detailed regional features, while pooling captures the image’s overall information.  In the cropping approach, we progressively crop the ViT patch features inward until reaching the central region, obtaining multi-scale features. These features are then arranged from left to right in the order of ``from central region to global". Cropping focuses on regional details, making it suitable for tasks requiring fine-grained image understanding. In contrast, the pooling method uses  adaptive pooling layers to generate multi-scale features that capture  varying levels of abstraction, which are then arranged from left to right in the order of ``from abstract to specific". This strategy is inspired by the hierarchical manner in which humans perceive or create images~\cite{autogressiveimage}, first capturing the global structure and then focusing on local details. Pooling is especially beneficial for tasks that require a more general understanding of the image. For both methods, the ViT patch features are reshaped to their original 2D shapes. They are then reorganized according to the ``from central region to global" or ``from abstract to specific" strategy, resulting in structured multi-scale features. Finally, convolutional kernels of varying sizes are applied to these multi-scale features to capture visual spatial tokens, which are then concatenated with the original visual tokens to form a comprehensive visual representation. 2) The DFI further enhances visual spatial features through a cross-attention mechanism. Without increasing the number of visual spatial tokens extracted by SFE, DFI derives fine-grained features from the large-size visual feature maps and integrates them into the visual spatial tokens to accomplish feature fusion, which further enhances the visual representation and thereby improves the detailed understanding ability of MLLMs.

In summary, our main contributions are as follows:
\begin{itemize}
\item \textbf{Visual spatial tokens enhance the visual representation of MLLMs.} We propose a novel Projector to capture visual spatial tokens, effectively extracting the spatial information among local adjacent ViT patch features.
\item \textbf{Two model variants handle diverse tasks.} LLaVA-SP-Cropping focuses on detailed features, while LLaVA-SP-Pooling captures global semantics, handling fine-grained and general visual understanding tasks respectively. 
\item \textbf{Performance improvements on various multimodal benchmarks.} \cref{eval}  demonstrates that LLaVA-SP finetuned with LoRA~\cite{lora} outperform LLaVA-1.5 on various multimodal benchmarks.
\end{itemize}


\section{Related Work}
\label{sec:formatting}
With the remarkable success of commercial MLLMs like OpenAI GPT-4V~\cite{gpt4v} and Google Gemini~\cite{gemini}, AI applications~\cite{vqa0,vqa1,vqa2} for text-image understanding have become a part of our daily lives. This development has sparked enthusiastic research among scholars on the visual language understanding capabilities of open-source MLLMs.

\subsection{Multimodal Large Language Models}

Research in multimodal large language models has focused on aligning visual and linguistic representation to improve interaction between the two domains. Flamingo~\cite{flamingo} introduced the Perceiver Resampler, which employed a cross-attention mechanism to integrate visual data into large language models (LLMs). 
The BLIP~\cite{blip,blip2,blip3,instructblip} and Qwen-VL~\cite{Qwen-VL,Qwen2VL} series developed the Q-former structure for visual-language alignment, using learnable parameter queries to compress visual information and reduce the number of visual tokens. Alternatively, the Mini-GPT4~\cite{minigpt4} and LLaVA series~\cite{llava,llava1.5} adopted a simple multilayer perceptron (MLP) as projector to map visual features into the language representation space of LLMs.

Furthermore, MLLMs such as VILA~\cite{vila}, MMICL~\cite{mmicl}, and MANTIS~\cite{mantis} have emphasized enhancing the quality of training data. These studies demonstrate that interleaved image-text datasets can better stimulate MLLMs' potential and improve contextual learning. Bunny-3B~\cite{bunny} leveraged an efficient data clustering compression technique to construct a high-quality dataset. Share-GPT4V~\cite{sharegpt4v} produced a detailed image-text description dataset using GPT-4V.

End-to-end MLLMs represent a cutting-edge area of research, focusing on direct processing of visual inputs without relying on pre-trained vision encoder. Fuyu-8B~\cite{fuyu}, EVE~\cite{eve}, SOLO~\cite{SOLO}, and OtterHD~\cite{otterhd} forgo pre-trained vision encoder and directly segment images into patches for input into LLMs instead. These methods allow MLLMs to bypass the limitations imposed by the prior knowledge of vision encoder, facilitating the learning of unaltered visual information. Our work builds on LLaVA-1.5, investigating the potential of vision encoder to enhance visual representation for MLLMs.

\subsection{Visual-Enhanced MLLMs}

Recent research in visual-enhanced MLLMs has concentrated on improving the visual component by increasing image resolution, fusing visual features, and designing efficient projectors. For example, LLaVA-HR~\cite{llavahd} introduced a mixture of resolution mechanism that combines information from low-resolution and high-resolution images. InternVL~\cite{internvl} developed a InternViT-6B model comparable in scale to LLM, enhancing its ability to process visual inputs. Additionally, InternVL1.5~\cite{internvl1.5}, LLaVA-NeXT~\cite{llavanext}, and LLaVA-UHD~\cite{llavauhd} implemented a dynamic resolution strategy to accommodate images of various aspect ratios, avoiding distortion caused by forced padding or resizing.

Studies on fusing visual features have produced notable advancements. Dense Connector~\cite{dense} fused features through methods such as sequential and channel concatenation, feature addition across different ViT layers. SPHINX~\cite{sphinx} integrated visual features from models like CLIP-ViT~\cite{clip}, ConvNext~\cite{convnet}, and DINOv2-ViT~\cite{dinov2} to extract diverse types of visual information. EAGLE~\cite{eagle} studied the impact of deformable attention fusion~\cite{deformable} on model performance. However, these techniques typically necessitate an increased number of tokens, which can lead to inefficiencies in both training and inference.

Some research efforts have specifically targeted the improvement of projector. Honeybee~\cite{honeybee} designed a Q-former structure projector based on convolutional neural network (CNN) and deformable attention~\cite{deformable} to enhance visual local information. DeCo~\cite{deco} applied adaptive average pooling layers to reduce the number of visual tokens and demonstrated its superiority over the Q-former.

Our work contributes to visual-language feature alignment, similar to Honeybee~\cite{honeybee}, by focusing on extracting spatial information from visual features.

\section{Methods}

\subsection{Overview}
The LLaVA-SP follows the design of LLaVA-1.5~\cite{llava1.5}, consisting of three parts: Vision Encoder, Projector, and LLM, as shown in \cref{achichect}:

\begin{figure}[t]
  \centering
\includegraphics[width=1.0\linewidth]{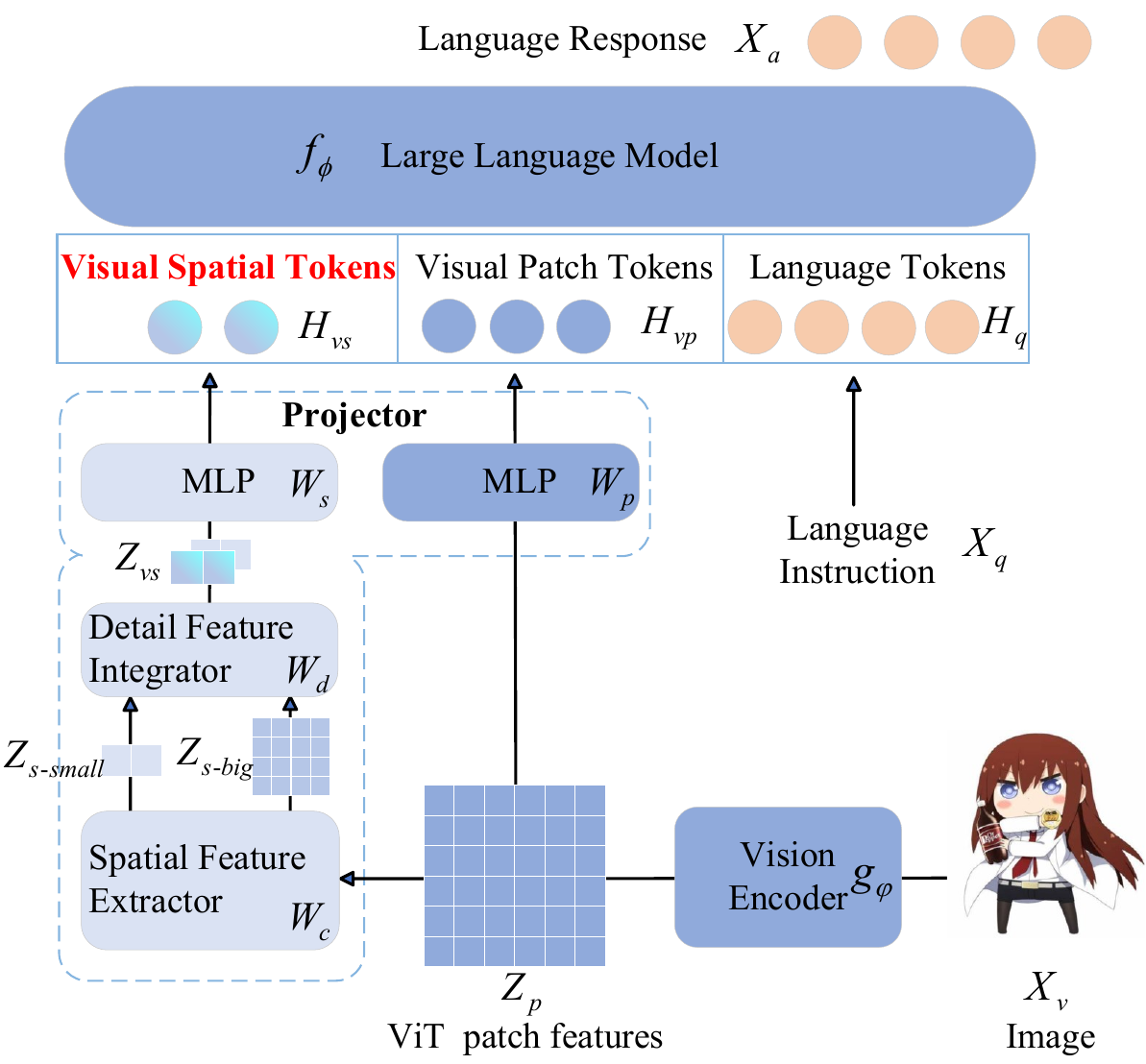}
   \caption{The architecture of LLaVA-SP is based on the structure of LLaVA-1.5~\cite{llava1.5}. The projector features two parallel branches, with the left branch dedicated to extracting visual spatial tokens.}
   \label{achichect}
   
\end{figure}

\noindent\textbf{Vision Encoder.} We employ the pre-trained CLIP-ViT-L/14-336 model~\cite{clip} as our vision encoder, denoted by $g _\varphi(\cdot)$, where $\varphi$ represents its parameters. When an image $X_v$ is provided as input, the encoder extracts ViT patch features, resulting in $Z_p = g(X_v)$. 

\noindent\textbf{Projector.} The projector maps visual features into the language representation space of the large language model. It consists of three components: SFE (trainable convolution matrices $W_c$), DFI (trainable linear matrices $W_d$), and two parallel MLPs ($W_s$ and $W_p$).
SFE begins the process by extracting visual spatial features $Z_s$ from ViT patch features $Z_p$. DFI mines fine-grained features by integrating small-scale ($Z_{s\text{-}small}$) and
large-scale ($Z_{s\text{-}big}$) features, further enriching the details of the visual spatial features $Z_{vs}$. The two parallel MLPs perform specialized transformations: $W_s$ converts spatial features $Z_{vs}$ into visual spatial tokens $H_{vs}$, while $W_p$ transforms ViT patch features $Z_p$ into visual patch tokens $H_{vp}$. This dual mapping ensures that distinct visual features are independently processed, preserving personalized information and aligning them within a consistent representation space.

\noindent\textbf{LLM.} We select Vicuna-1.5~\cite{vicuna} as the LLM. The language instruction is represented as language tokens $H_q$ through the LLM embedding layer. As depicted in \cref{achichect}, $H_{vs}$, $H_{vp}$, and $H_q$ are concatenated sequentially and input into the LLM for autoregressive training. The formula for calculating the prediction probability $p$ of the next token at the current position $i$ is expressed as follows:
\begin{equation}
  p(X_a|X_v,X_q) = \prod_i^L p(X_i|X_v,X_{q,<i},X_{a,<i}),
  \label{eq:important}
\end{equation}
where $L$ is the length of the input sequence, $X_a$ is the answer, $X_q$ is the query, $X_v$ is the image, and $X_{<i}$ refers to the sequence of tokens preceding the current token $X_i$.
\begin{figure*}
  \centering
  \begin{subfigure}{0.497\linewidth}
    \includegraphics[width=1.0\linewidth]{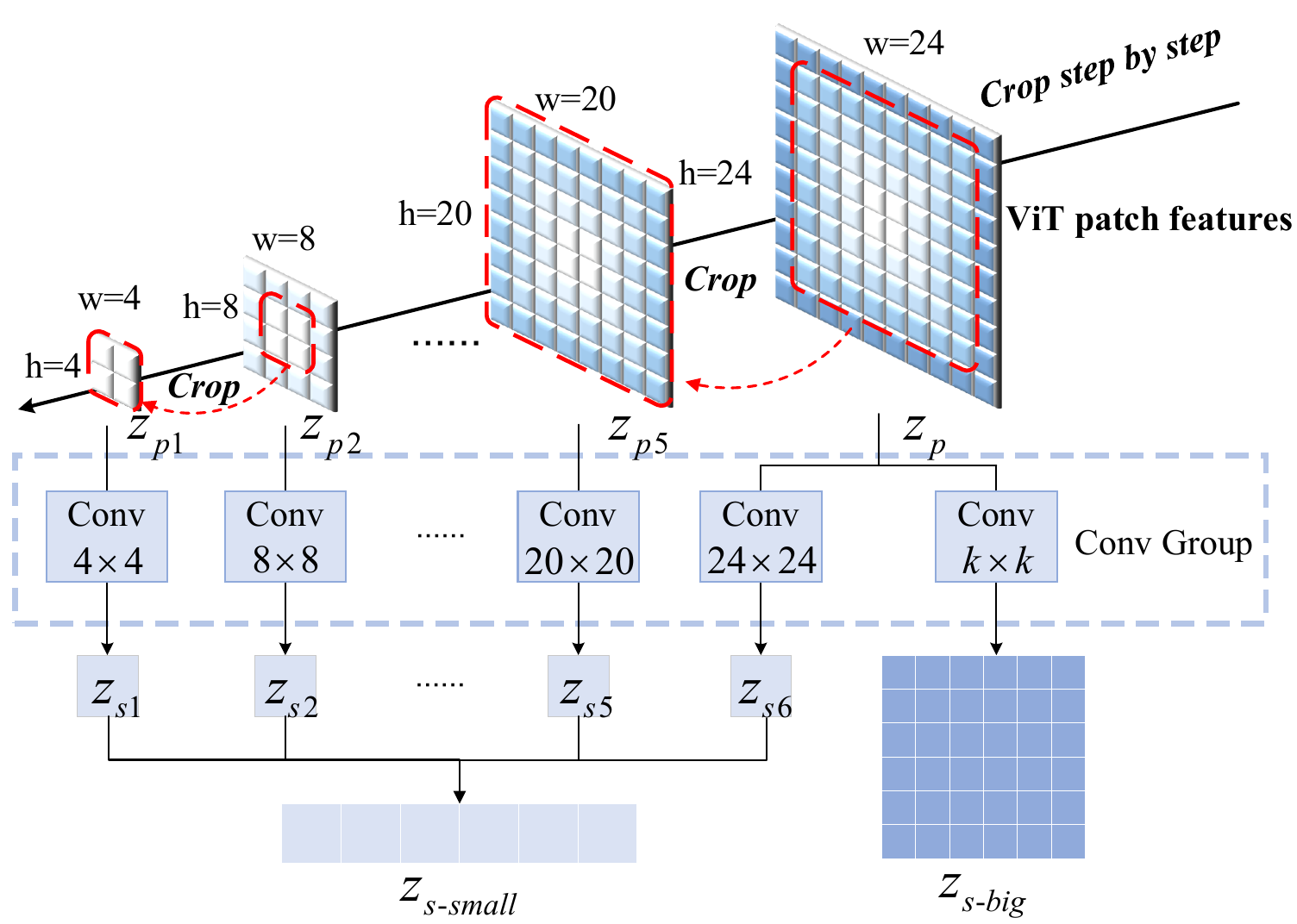}
    \caption{The SFE module in LLaVA-SP-Cropping model.}
    \label{fig:crop}
  \end{subfigure}
  \hfill
  \begin{subfigure}{0.497\linewidth}
    \includegraphics[width=1.0\linewidth]{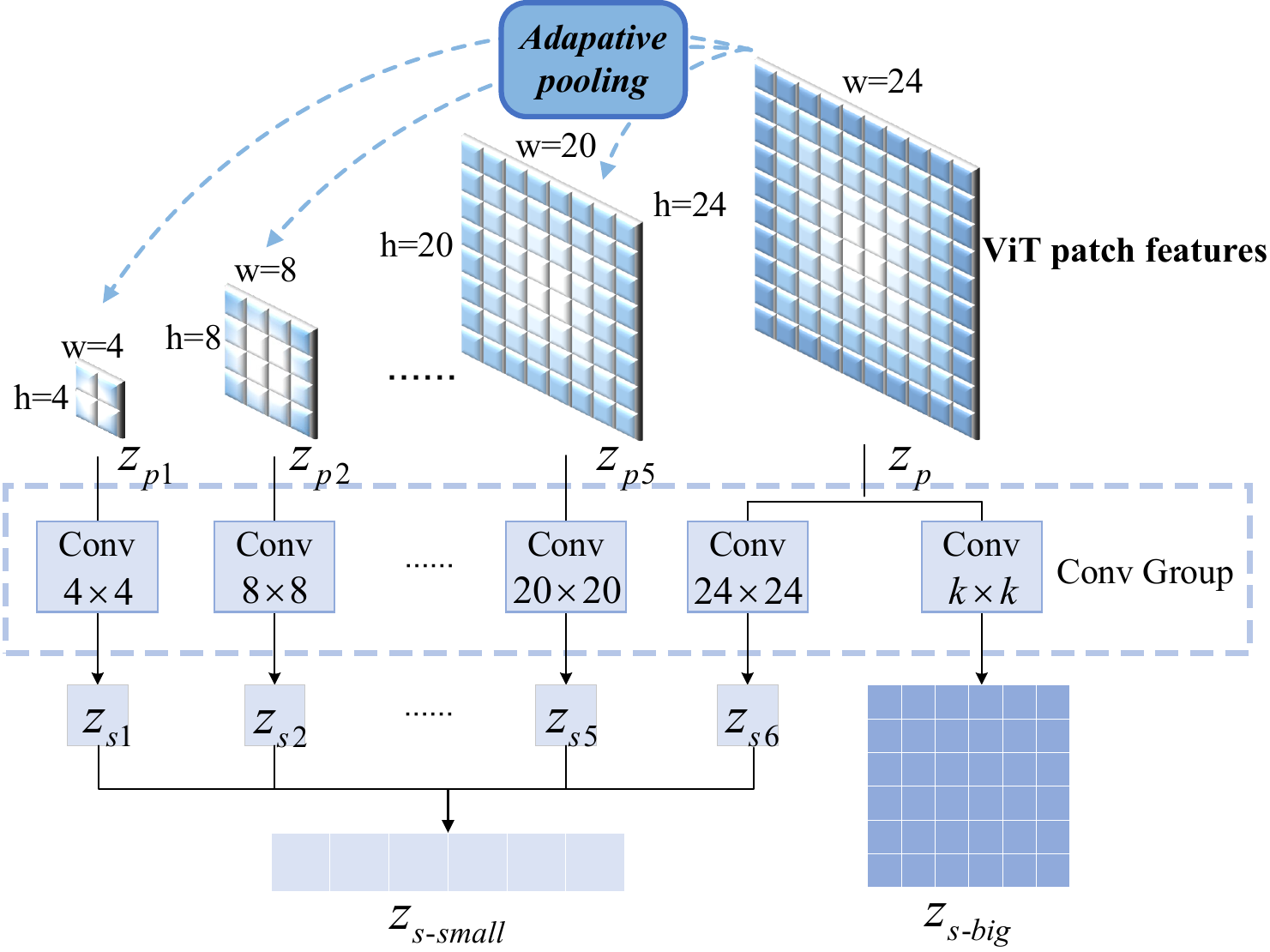}
    \caption{The SFE module in LLaVA-SP-Pooling model.}
    \label{fig:pooling}
  \end{subfigure}
  \caption{\textbf{SFE Structure.} (a) illustrates the process of obtaining precise multi-scale features using the cropping operation, simulating the arrangement of visual spatial features as ``from central region to global", emphasizing details in image regions. (b) demonstrates the method of obtaining abstract feature maps at multi-scale using adaptive pooling, simulating the arrangement of visual spatial features ``from abstract to specific", emphasizing the global semantics of the image. We use a group of convolutional kernels to extract visual spatial features $Z_{s\text{-}small}$, and $Z_{s\text{-}big}$ is used to feature fusion in DFI.}
  \label{fig:sfe}
\end{figure*}
\subsection{Spatial Feature Extractor}
\label{sec:sfe}
Traditional visual tokens are arranged in a 1D manner, from left to right and top to bottom, which disrupts the original 2D spatial relationships of the visual features and causes information confusion. Therefore, we propose the Spatial Feature Extractor (SFE) to capture the spatial relational information of visual features, serving as supplements to the original visual representation. The design of SFE follows two principles: 1) Obtaining multi-scale features that capture the 2D spatial structure of image. 2) Using convolutional kernels to extract visual spatial features.

To obtain the multi-scale features, we can operate on ViT patch features using cropping or pooling. 

\noindent\textbf{Cropping.} \cref{fig:crop} shows that obtain multi-scale features by cropping. SFE rearranges CLIP-ViT-L/14-336 patch features to their original 2D shape $Z_p \in \mathbb{R}^{\sqrt{N} \times \sqrt{N} \times C}$, where $N = 576$ represents the number of visual patches and $C$ denotes the feature dimension. In the first step, we obtain all ViT patch features $Z_{p6} = Z_{p} \in \mathbb{R}^{24 \times 24 \times C}$. In the second step, using $Z_{p6}$ as the reference, we crop inward with a stride = 2 to obtain $Z_{p5} \in \mathbb{R}^{20 \times 20 \times C}$. This feature cropping process is repeated until the remaining central region features are too small to crop, like $Z_{p1} \in \mathbb{R}^{4 \times 4 \times C}$ in \cref{fig:crop}. This process generates multi-scale features $(Z_{p1},Z_{p2},Z_{p3},Z_{p4},Z_{p5},Z_{p6})$ arranged in ``from central region to global", emphasizing details in the image regions.

\noindent\textbf{Pooling.} \cref{fig:pooling} shows that obtain multi-scale features by pooling. SFE uses adaptive average pooling~\cite{adapool} to simulate the process of visual perception and creation from abstract to specific. Smaller feature maps lose more information and represent more abstract information, while larger feature maps convey more concrete details. The multi-scale feature sequence is arranged in ``from abstract to specific", emphasizing image global semantics.

Next, we utilize the inherent spatial modeling capability of convolution to extract spatial features. Convolutional kernels with sizes $k = 4,8,12,16,20,24$ can fully cover $(Z_{p1},Z_{p2},Z_{p3},Z_{p4},Z_{p5},Z_{p6})$ and compute visual spatial features $Z_{s\text{-}small}$ through concatenation in sequence dimension: 
\begin{equation}
 Z_{si}=conv_k(Z_{pi; k = 4i, i=1,2,3,4,5,6}),
  \label{eq:zs}
\end{equation}
\begin{equation}
Z_{s\text{-}small}=concat(Z_{s1},Z_{s2},Z_{s3},Z_{s4},Z_{s5},Z_{s6}),
  \label{eq:zs-small}
\end{equation}
where  
 $Z_{si}\in\mathbb{R}^{1 \times 1 \times C}$, $Z_{s\text{-}small}\in\mathbb{R}^{6 \times 1 \times C}$, $concat$ denotes concatenation and $conv$ denotes convolutional kernel.

\begin{figure}[t]
  \centering
  
\includegraphics[width=1.0\linewidth]{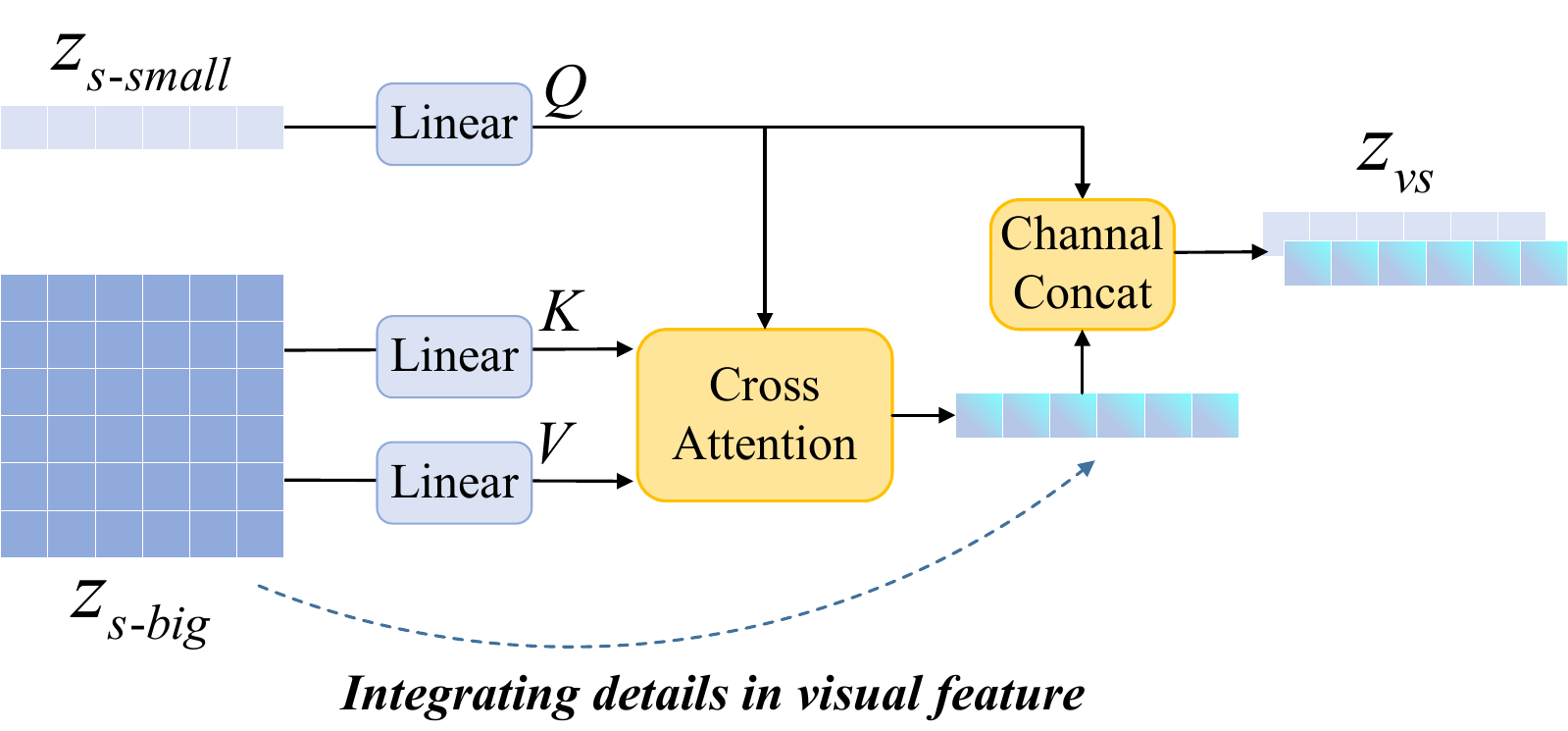}
   \caption{\textbf{DFI architecture.} Integrating $Z_{s\text{-}big}$ details and injecting them into $Z_{s\text{-}small}$.}
   \label{fuse}
   
\end{figure}


\begin{table*}[t!]

\footnotesize
\centering
\setlength{\tabcolsep}{4.55pt}
\begin{tabular*}{1.0\linewidth}{l c c|c c c c c|c c c c c c}
\toprule
Method & LLM & Res. & VQA\textsuperscript{v2} & GQA & VizWiz & SQA\textsuperscript{I} & VQA\textsuperscript{T} & POPE & MME\textsuperscript{P} & MMB & SEED\textsuperscript{I} & LLaVA\textsuperscript{W} & MM-Vet\\
\midrule
BLIP-2~\cite{blip2} & Vicuna-13B & 224 & 41.0 & 41.0 & 19.6 & 61.0 & 42.5 & 85.3 & 1293.8 & – & 46.4 & 38.1 & 22.4 \\

InstructBLIP~\cite{instructblip} & Vicuna-7B & 224 & – & 49.2 & 34.5 & 60.5 & 50.1 & – & – & 36.0 & 53.4 & 60.9 & 26.2    \\

InstructBLIP~\cite{instructblip} & Vicuna-13B & 224 & – &
49.5 & 33.4 & 63.1 & 50.7 & 78.9 & 1212.8 & – & – & 58.2 & 25.6 \\

Shikra~\cite{shikra} & Vicuna-13B & 224 & 77.4 & – & – & – & – & – & – & 58.8 & – & – & –
\\

Qwen-VL~\cite{Qwen-VL} & Qwen-7B & 448 & 78.8 & 59.3 & 35.2 & 67.1 & \textbf{63.8} & – & – & 38.2 & 56.3 & – & –    \\

Qwen-VL-Chat~\cite{Qwen-VL} & Qwen-7B & 448 & 78.2 & 57.5 & 38.9 & 68.2 & \underline{61.5} & – & \underline{1487.5} & 60.6 & 58.2 & – & – \\

DeCo~\cite{deco} & Vicuna-7B & 336 & 74.0 & 54.1 & 49.7 & – & 56.2 & 85.9 & 1373.4 & 60.6 & 62.8 & – & – \\

LLaVA-1.5\dag~\cite{llava1.5} & Vicuna-7B & 336 & 78.5 & 62.0 & 50.0 & 66.8 & 58.2 & 85.9 & \textbf{1510.7} & 64.3 & 66.2 & 63.4 & 30.5 \\
LLaVA-1.5*~\cite{llava1.5} & Vicuna-7B & 336 & 78.4 & 61.9 & 45.7 & 67.6 & 56.2 & 85.8 & 1477.4 & 64.5 & 67.0 & 64.2 & 32.1 \\

\midrule
\rowcolor{gray!20}
LLaVA-SP-Cropping & Vicuna-7B & 336 & \underline{79.2} & \underline{62.4} & \underline{50.1} & \textbf{69.7} & 58.7 & \underline{86.4} & 1473.8 & \textbf{65.8} & \textbf{67.6} & \underline{66.7} & \underline{32.2} \\

\rowcolor{gray!20}
LLaVA-SP-Pooling & Vicuna-7B & 336 & 79.1 & \textbf{62.5} & \textbf{51.6} & \underline{69.0} & 58.3 & \textbf{86.5} & 1475.9 & \underline{65.7} & \underline{67.5} & \textbf{68.3} & \textbf{33.4} \\

\bottomrule
\end{tabular*}

\caption{\textbf{Comparison with SoTA methods on 11 benchmarks}. The two versions of LLaVA-SP \textbf{fine-tuned with LoRA} surpassed LLaVA-1.5 on 10 / 11 benchmarks.
* indicates reproduced results
using LoRA while † denotes the full-training results reported in LLaVA-1.5~\cite{llava1.5}, and Res. indicates input image resolution. The best and second-best results are \textbf{bolded} and \underline{underlined}, respectively.}
\label{table:1}
\end{table*}


\subsection{Detail Feature Integrator}
\label{sec:dfi}
Our goal in designing DFI was to address the trade-off in SFE, where large convolution kernels capture a broad receptive field but miss finer details, while smaller kernels increase token count. To avoid increasing visual spatial tokens, and thus prevent the training and inference costs associated with long input sequences to LLM. DFI uses an attention mechanism to inject fine-grained features from smaller convolution kernels into the six tokens generated by SFE. 

Mentioned in ~\cref{sec:sfe}, $Z_{s\text{-}small}$ represents six visual spatial features. $Z_{s\text{-}big}$ is a feature map extracted using smaller kernels (the deep blue kernel on the far right of Conv Group in \cref{fig:crop,fig:pooling}). As shown in \cref{fuse}: $Z_{s\text{-}small}$ is used as the query, while $Z_{s\text{-}big}$ serves as the key and value. Through the cross-attention mechanism, fine-grained features are mined from $Z_{s\text{-}big}$ and injected into $Z_{s\text{-}small}$. Then we $concat$ attention features and  $Z_{s\text{-}small}$ in channel dimension, extracting visual spatial features $Z_{vs}$:
\begin{equation}
Z_{vs}=concat([Z_{s\text{-}small},softmax(\frac{Q\times K^\top}{\sqrt{d_k}})\times V]),
\label{eq:attention}
\end{equation}

\noindent where $Z_{vs}\in\mathbb{R}^{6 \times 1 \times 2C}$, $d_k$ is feature dimension, $Q=W_Q(Z_{s\text{-}small})$, $K=W_K(Z_{s\text{-}big})$, $V=W_V(Z_{s\text{-}big})$. $W_Q$, $W_K$ and $W_V$ are trainable linear matrices.

\section{Experiments}

\subsection{Setting}

LLaVA-SP is built on LLaVA-1.5~\cite{llava1.5}, including the same model components, training datasets and two-stage training strategy. Using CLIP-ViT-L/14-336~\cite{clip} as the vision encoder, and Vicuna-1.5-7B~\cite{vicuna} as the LLM. The training dataset includes 558K pre-training data~\cite{llava1.5} (sourced from LAION~\cite{laion}, Conceptual Captions~\cite{cc12m}, and SBU Captions~\cite{sbu}) and 665K instruction-following data (containing LLaVA Synthetic Data~\cite{llava1.5}). The two-stage training strategy includes pre-training and fine-tuning, and \textbf{we fine-tune the LLM using LoRA~\cite{lora} in all of our experiments.} We conducted performance evaluations on various benchmarks, consisting of: 1) General visual question answering, like VQAv2 (VQA\textsuperscript{v2})~\cite{vqav2}, TextVQA (VQA)\textsuperscript{T})~\cite{textvqa}, ScienceQA-Image (SQA)~\cite{sqa}, GQA~\cite{gqa} and Vizwiz~\cite{vizwiz}. 2) Comprehensive benchmarks, like MM-Vet~\cite{mmvet}, MMBench (MMB)~\cite{mmbench}, LLaVA-Bench-In-the-Wild (LLaVA\textsuperscript{W})~\cite{llava1.5},  MME-Perception (MME\textsuperscript{P})~\cite{mme} and SEED-Bench-Image (SEED\textsuperscript{I})~\cite{li2023seed}.
3) Hallucination benchmark like POPE~\cite{pope} and MMVP~\cite{mmvp}. 4) Visual ground benchmark RefCOCO~\cite{refcoco}. LLaVA-Bench and MM-Vet score is reported by GPT-4-0613.

\begin{table}[t]
    \centering
    \small
    \setlength{\tabcolsep}{1.8pt}
    \begin{tabular}{l c p{2cm} c| c}
    \toprule
    Method & LLM  & Vision Encoder & N & Tokens / s  \\
    
\midrule
Qwen-VL~\cite{Qwen-VL}  &  7B & CLIP-ViT-G  & 256 & 13.01 \\

Qwen2-VL~\cite{Qwen2VL} &  7B & ViT-B & dynamic & 12.23 \\

LLaVA-1.5~\cite{llava1.5} &  7B  & CLIP-ViT-L  & 576 &   20.76   \\
LLaVA-SP-Cropping &  7B & CLIP-ViT-L  & 582 &   20.51   \\
LLaVA-SP-Pooling &  7B & CLIP-ViT-L  & 582 &   20.28   \\
   
    \bottomrule
    \end{tabular}
    \caption{\textbf{Inference speed evaluation.} ``N" represents the number of visual tokens. More visual tokens lead to longer runtime. Runtime of LLaVA-SP is comparable to LLaVA-1.5.}
    \label{tab:speed}
\end{table}


\begin{table*}[t!]

\footnotesize
\centering
\setlength{\tabcolsep}{4.8pt} 
\begin{tabular*}{1.0\linewidth}{l c c| c c c c c | c  c c c c c c c }
\toprule

Method & Version & Type & VQA\textsuperscript{v2} & GQA & VizWiz & SQA\textsuperscript{I} & VQA\textsuperscript{T} & POPE & MME\textsuperscript{P} & MMB & SEED\textsuperscript{I} & LLaVA\textsuperscript{W} & MM-Vet & Avg\textsuperscript{N}\\
\midrule

LLaVA-1.5* & - & - & 78.4 & 61.9 & 45.7 & 67.6 & 56.2 & 85.8 & 1477.4 & 64.5 & 67.0 & 64.2 & 32.1 & 63.4 \\
\midrule

\ \ +SFE &Cropping		&T	&79.1	&61.8	&53.0	&69.4	&58.0	&87.0	&1478.2	&65.0	&67.0	&64.7	&30.6	&64.5 \\

\rowcolor{gray!20}
\ \ +SFE &Cropping		&C	&\textbf{79.1}	&\textbf{62.7}	&49.7	&68.7	&\textbf{58.3}	&86.1	&1461.5	&\textbf{65.8}	&\textbf{67.1}	&\textbf{66.6}	&\textbf{33.6}	&\textbf{64.6} \\

\midrule

\ \ +SFE & Pooling		&T	&79.1	&62.8	&47.4	&69.2	&57.6	&86.5	&1475.9	&66.5	&67.7	&67.4	&30.0	&64.4 \\

\rowcolor{gray!20}
\ \ +SFE &Pooling		&C	&\textbf{79.2}	&62.7	&\textbf{49.6}	&\textbf{70.0}	&\textbf{58.5}	&\textbf{86.8}	&1474.8	&66.3	&\textbf{67.9}	&\textbf{67.5}	&\textbf{32.1}	&\textbf{64.9} \\

\bottomrule
\end{tabular*}
\caption{\textbf{Ablation: Convolution vs. Transformer blocks.} ``Type" represents the model structure type used by SFE, ``C"denotes convolutional kernels and ``T" denotes tranformer blocks. * indicates reproduced results using LoRA. 
Experiments show that convolution has better performance than transformer blocks.}
\label{table:2}
\end{table*}


\begin{table*}[t!]

\footnotesize
\centering
\setlength{\tabcolsep}{5.435pt}
\begin{tabular*}{1.0\linewidth}{l c c  | c c c c c | c  c c c c c c  }
\toprule

Feature shape & S & N & VQA\textsuperscript{v2} & GQA & VizWiz & SQA\textsuperscript{I} & VQA\textsuperscript{T} & POPE & MME\textsuperscript{P} & MMB & SEED\textsuperscript{I} & LLaVA\textsuperscript{W} & MM-Vet & Avg\textsuperscript{N}\\
\midrule

(24) & - & 1 & 79.0 & 62.5 & 47.9 & 68.2 & 57.9 & 86.1 & 1473.4 & 65.2 & 66.4 & 66.2 & 28.6 & 63.4 \\

(8,16,24) & 4	&3	& 79.1 & 62.6	& 47.2	& 69.3 & 58.2	&86.2 	&1454.3	&64.5&	67.5 &66.4	&31.9	& 64.1\\

\rowcolor{gray!20}
(4,8,12,16,20,24) & 2  &6	&\textbf{79.1}	&\textbf{62.7}	&\textbf{49.7}	&68.7	&\underline{58.3}	&86.1&\underline{1461.5}	&\underline{65.8}	&\underline{67.1}	&\underline{66.6}	&\textbf{33.6}	&\textbf{64.6} \\

(2,4,6...20,22,24) & 1	& 12 &  79.0 &62.6	&46.9	&70.1	&58.1	&86.7	&1450.2	&67.2	&66.8	&67.9	&32.8	& 64.6\\

\bottomrule
\end{tabular*}
\caption{\textbf{Ablation: The number of visual spatial tokens}. ``S" represents the step size by which each cropping
reduces inward, ``N" represents the number of visual spatial tokens, and ``Feature shape" represents the shape of multi-scale features $(Z_{p1},Z_{p2},Z_{p3},Z_{p4},Z_{p5},Z_{p6})$ when $N=6$. Experiments show that six visual spatial tokens can effectively capture spatial information from ViT patch features.}
\label{table:num}
\end{table*}


\subsection{Main Results}
\cref{table:1} shows the evaluation results across 11 benchmarks. Both LLaVA-SP-Cropping and LLaVA-SP-Pooling demonstrate significant performance improvements on 10 out of 11 benchmarks, compared to LLaVA-1.5* reproducd using LoRA. Our best model achieves the following improvements: VQAv2 by +0.8\%, GQA by +0.6\%, VisWiz by +5.9\%, SQA-IMG by +2.1\%, TextVQA by +2.5\%, POPE by +0.7\%, MMBench by +1.3\%, SEED-IMG by +0.6\%, LLaVA-Bench by +1.3\%, and MM-Vet by +1.3\%. We also report the max-normalized average score Avg\textsuperscript{N}~\cite{avg1, honeybee} across 11 benchmarks, where LLaVA-SP-Cropping and LLaVA-SP-Pooling, fine-tuned with LoRA, improved by 1.5\% and 1.6\%, respectively, over the fully trained  LLaVA-1.5~\cite{llava1.5}.

We evaluated the model's inference speed on a single A40 GPU, with all LLMs being 7B parameters to ensure a fair evaluation. \cref{tab:speed} shows the inference speed of LLaVA-SP-Cropping and LLaVA-SP-Pooling is 20.51 and 20.28 tokens per second, respectively, which are comparable to LLaVA-1.5 and faster than methods using larger ViT or dynamic visual tokens.


\subsection{Analysis of Spatial Feature Extractor}

\textbf{Overall.} \cref{table:2} indicates that the average scores for LLaVA +SFE-Cropping and LLaVA +SFE-Pooling in  Avg\textsuperscript{N} were 64.6 and 64.9, respectively, representing improvements of 1.2\% and 1.5\% over LLaVA-1.5*. These results confirm the effectiveness of the SFE. Additionally, \cref{table:2} shows pooling method achieves 0.3\% higher Avg\textsuperscript{N} than cropping method on general VQA benchmarks. This is because pooling better captures overall information of images. In contrast, cropping is better at handling fine-grained image understanding tasks, which we discuss in  ~\cref{sec:vg}.

\begin{table}[t]

\footnotesize
\centering
\setlength{\tabcolsep}{4pt}
    \begin{tabular}{l|c|ccc|cc|c}
        \toprule
        Method &\multicolumn{1}{c|}{MME} & \multicolumn{3}{c|}{MMB} & \multicolumn{2}{c|}{SEED} & Avg\textsuperscript{N} \\
         & POS & SR & OL & PR & SR & IL  & \\
        \midrule

     LLaVA-1.5 Baseline 

     &128.3  & 20.0 & 44.4 & 25.0 & 51.1 & 59.9 & 44.1 \\
     
      \ \   + Sliding window  & 127.2 & 19.8 & 47.3 & 27.4 & 50.1 & 60.2 & 44.7 \\
\rowcolor{gray!20}
\ \ + SP-Cropping  & 126.7 & 24.4 & 50.6 & 29.2 & 49.8 & 61.7 & 46.5 \\
        \bottomrule
    \end{tabular}
    \caption{Performance comparison of token design in SFE.  }
    \label{tab:sfe}
\end{table}

\noindent\textbf{Ablation: Token design in SFE.} \cref{tab:sfe} compares the performance of SP-Cropping and the token design using Sliding window, where we crop features of the same size from top to bottom and left to right, then use convolutional kernels to extract spatial tokens. However, sliding window tokens disrupt 2D spatial relationships. In contrast, LLaVA-SP integrates human visual perception, considering adjacent features in all directions, making it more effective.

\noindent\textbf{Ablation: Convolution vs. Transformer blocks.} \cref{table:2} compares the performance of the SFE module with convolution and transformer blocks. For both LLaVA +SFE-Cropping and LLaVA +SFE-Pooling, the convolution outperforms the transformer blocks. The convolution-based model improves performance across all 10 benchmarks, while the transformer-based model shows weaker performance on GQA, SEED-IMG, and MM-Vet. This can be attributed to the fact that convolution excels at extracting spatial information from images, as validated by our experiments. Thus, SFE uses convolution in all our experiments. We also tried multi-layer CNN blocks and small-scale convolutional kernels, but both caused training crashes. The multi-layer CNN blocks likely caused gradient explosion or vanishing, resulting in some parameters becoming excessively large. Additionally, small-scale convolutional kernels cannot cover all the feature maps, which results in the generation of more visual spatial tokens. When concatenated with ViT patch tokens, this leads to feature confusion.

\begin{table*}[t]

\footnotesize
\centering
\setlength{\tabcolsep}{2.88pt} 

\begin{tabular*}{1.0\linewidth}
{lccc|ccccc|ccccccc}

\toprule

Method & Version &$Z_{s\text{-}big}$
size & Conv size & VQA\textsuperscript{v2} & GQA & VizWiz & SQA\textsuperscript{I} & VQA\textsuperscript{T} & POPE & MME\textsuperscript{P} & MMB & SEED\textsuperscript{I} & LLaVA\textsuperscript{W} & MM-Vet & Avg\textsuperscript{N} \\
\midrule

LLaVA-1.5* &	- &	- &	- &	78.4 &	61.9 &	45.7  &	67.6 &	56.2 &	85.8 &	1477.4 &	64.5 &	67.0 &	64.2 &	32.1 &	63.4  \\

\midrule

\ \ +SFE & Cropping&	- &	- &	79.1 &	62.7 &	49.7 &	68.7 &	58.3 &	86.1 &	1461.5 &	65.8 &	67.1 &	66.6 &	33.6 &	64.6  \\

\ \ +SFE+DFI &	Cropping&$11\times11$ &	$4\times4$ &	{\color{red} 79.3} &
62.7 &
{\color{ForestGreen} 49.3} &
{\color{red} 68.9} &
{\color{red} 58.5} &
{\color{red} 86.2} &
{\color{red} 1467.2} &
{\color{red} 66.6} &
{\color{red} 67.6} &
{\color{ForestGreen} 65.3} &
{\color{red} 34.4} &
{\color{red} 64.7} \\

\ \ +SFE+DFI &	Cropping&$9\times9$ &	$8\times8$ &	{\color{red} 79.2} &
{\color{red} 62.8} &
{\color{ForestGreen} 48.0} &
{\color{red} 70.2} &
{\color{red} 58.7} &
{\color{red} 86.2} &
{\color{ForestGreen} 1461.4} &	{\color{ForestGreen} 65.5} &
{\color{red} 67.2} &
{\color{ForestGreen} 64.6} &
{\color{ForestGreen} 30.3} &
{\color{ForestGreen} 64.2} \\

\ \ +SFE+DFI &	Cropping&$7\times7$ &	$12\times12$ & {\color{red} 79.2} &
62.7 &
{\color{ForestGreen} 48.5} &
{\color{red} 70.8} &
{\color{red} 58.7} &
{\color{red} 86.7} &
{\color{red} 1490.4} &
{\color{red} 66.3} &
{\color{ForestGreen} 66.9} &
{\color{ForestGreen} 64.7} &
{\color{ForestGreen} 33.1} &
{\color{red} 64.7}  \\

\rowcolor{gray!20}
\ \ +SFE+DFI &	Cropping&$5\times5$ &	$16\times16$ & {\color{red} 79.2} &	
{\color{ForestGreen} 62.4} &	
{\color{red} 50.1} &
{\color{red} 69.7} &
{\color{red} 58.7} &
{\color{red} 86.4} &
{\color{red} 1473.8} &
65.8 &	
{\color{red} 67.6} &
{\color{red} 66.7} &
{\color{ForestGreen} 32.2} &
{\color{red} 64.8}  \\

\midrule

\ \ +SFE & Pooling &- & - &	 	79.2 &	62.7 &	49.6 &	70.0 &	58.5 &	86.8 &	1474.8 &	66.3 &	67.9 &	67.5 &	32.1 &	64.9 \\

\ \ +SFE+DFI & 	Pooling&$11\times11$
&	$4\times4$ &	79.2 &
{\color{red} 62.9} &
{\color{red} 50.7} &
{\color{ForestGreen} 69.4} &
{\color{ForestGreen} 58.4}  &
{\color{ForestGreen} 86.6}&
{\color{ForestGreen} 1466.4} &
{\color{ForestGreen} 65.9} &
{\color{ForestGreen} 67.7} &
{\color{ForestGreen} 65.1} &
{\color{red} 32.2} &
{\color{ForestGreen} 64.7} \\

\ \ +SFE+DFI &	Pooling&$9\times9$ &	$8\times8$ &	79.2 &
{\color{red} 62.9} &
{\color{ForestGreen} 49.0} &
{\color{red} 70.2} &
{\color{ForestGreen} 58.2} &
{\color{ForestGreen} 86.7} &
{\color{ForestGreen} 1457.7} &
{\color{red} 66.8} &
{\color{ForestGreen} 67.3} &
{\color{ForestGreen} 64.8} &
{\color{ForestGreen} 30.4} &
{\color{ForestGreen} 64.4} \\

\ \ +SFE+DFI &	Pooling&$7\times7$ &	$12\times12$ &	{\color{ForestGreen} 79.1} &
62.7 &
{\color{ForestGreen} 47.5} &
{\color{ForestGreen} 69.1} &
{\color{ForestGreen} 58.4} &
{\color{ForestGreen} 86.0} &
{\color{red} 1485.7} &
{\color{ForestGreen} 65.2} &
{\color{ForestGreen} 67.0} &
67.5 &
{\color{ForestGreen} 31.0} &
{\color{ForestGreen} 64.3} \\

\rowcolor{gray!20}
\ \ +SFE+DFI &	Pooling&$5\times5$ &	$16\times16$ &
{\color{ForestGreen} 79.1} &
{\color{ForestGreen} 62.5} &
{\color{red} 51.6} &
{\color{ForestGreen} 69.0} &
{\color{ForestGreen} 58.3} &
{\color{ForestGreen} 86.5} &
{\color{red} 1475.9} &
{\color{ForestGreen} 65.7} &
{\color{ForestGreen} 67.5} &
{\color{red} 68.3} &
{\color{red} 33.4} &
{\color{red} 65.1}  \\

\bottomrule
\end{tabular*}
\caption{\textbf{Ablation: $Z_{s\text{-}big}$ feature map size}. Experiments show that when $Z_{s\text{-}big} = 5\times5$, the improvements are most noticeable in both Cropping and Pooling methods. * indicates reproduced results using LoRA. 
Compared with LLaVA+SFE, performance {\color{red} increases} and {\color{ForestGreen} decreases} are marked in {\color{red} red} and {\color{ForestGreen} green}, respectively.}

\label{table:3}
\end{table*}

\noindent\textbf{Ablation: The number of visual spatial tokens.} We conducted experiments on LLaVA +SFE-Cropping to explore the impact of different quantities of visual spatial tokens on model performance. The ``crop step by step" process generates multi-scale features. To ensure consistent shape increments for these multi-scale features, the stride size of each inward cropping step affects the number of visual spatial tokens. \cref{table:num} shows that the model performs better with more tokens, with the best performance observed at 6 tokens.  Although the performance with 12 tokens is comparable to that with 6, it doubles the parameters and slows down inference speed. Using only six visual spatial tokens effectively captures the spatial information of ViT patch features.

\subsection{Analysis of Detail Feature Integrator}

\textbf{Ablation: $Z_{s\text{-}big}$ feature map size.} We investigate how different feature granularities affect visual feature fusion. The size $k$ of the convolutional kernel (the deep-blue-colored convolutional kernel on the most right in \cref{fig:crop,fig:pooling}) controls the size of the feature map $Z_{s\text{-}big}$. The kernel size $k$ was set to even numbers ($k$=16, $s$=2, $n$=25; $k$=12, $s$=2, $n$=49; $k$=8, $s$=2, $n$=81; $k$=4, $s$=2, $n$=121), where $s$ is the stride, and $n$ is the resulting feature length. specifically, the shape of ViT patch features encoded by CLIP-ViT-L/14-336 is $24\times24$. We use an even-sized convolutional kernel to ensure that the feature area remains consistent for each convolutional sliding window operation. In contrast, an odd-sized kernel requires padding feature map margin with 0 or 1, which disrupts the original visual features. As shown in \cref{table:3}, 
when
$Z_{s\text{-}big} = 5\times5$
yielded the best performance improvement, because the largest convolutional kernel, $16\times16$, capturing a broader range of visual spatial information.

\noindent\textbf{Deep analysis.} Interestingly, DFI enhances the performance of LLaVA-SP-Cropping while negatively affecting LLaVA-SP-Pooling. In LLaVA-SP-Pooling, although the $16 \times 16$ convolutional kernel improves the overall average score across the benchmarks, most individual benchmark scores still decline. This difference can be attributed to the distinct modeling of the six visual spatial tokens: LLaVA-SP-Cropping directly crops feature maps of various sizes from the ViT patch features, preserving the original feature details. In contrast, LLaVA-SP-Pooling applies adaptive average pooling to obtain the six feature maps, performing operations similar to low-pass filtering, which abstracts the original features.  Since the surrounding local feature values are similar, the attention mechanism struggles to focus on which specific feature point is more significant, ultimately impairing visual feature fusion.

 \begin{table*}[t]
\footnotesize
\centering
\setlength{\tabcolsep}{4.3pt} 

\begin{tabular}{lccc|ccc|cccccc}
\toprule
Method & Vision Encoder & LLM & Res. & GQA &  SQA\textsuperscript{I} & VQA\textsuperscript{T} & POPE & MME\textsuperscript{P} & MMB & SEED\textsuperscript{I}   & MM-Vet & Avg\textsuperscript{N}\\

\midrule
LLaVA-1.5 & SigLIP-L/16 & Vicuna-7B &384 & 61.3 &  66.4 & 57.6 & 85.1 & 1450.0 & 65.2 & 67.9  & \underline{32.2} & 63.5 \\

\rowcolor{gray!20}
LLaVA-SP-Cropping & SigLIP-L/16 &   Vicuna-7B &384 & \underline{62.4} &  \textbf{68.9} & \textbf{59.9} & \textbf{85.7} & \underline{1509.2} & \textbf{65.6} & \underline{68.0} & 31.9 & \underline{64.7} \\

\rowcolor{gray!20}
LLaVA-SP-Pooling & SigLIP-L/16  &
Vicuna-7B &384 &  \textbf{62.9}  & \underline{68.6} & \underline{59.6} & \textbf{85.7} & \textbf{1514.8} & \underline{65.5}  & \textbf{68.3}  & \textbf{33.3} & \textbf{65.0} \\

\midrule

InternVL-2.0~\cite{internvl1.5} & InternViT‑300M  &
Qwen2‑0.5B &448 &  56.8  & 56.7 & 41.2 & 84.6 & 1064.0 & 52.1  & 55.5 & 20.4 & 52.4 \\

\rowcolor{gray!20}
InternVL-2.0-SP-Cropping & InternViT‑300M  &
Qwen2‑0.5B &448 &  \textbf{58.4}  & \underline{57.9} & \underline{41.4} & \textbf{85.2} & \underline{1187.5} & \textbf{53.3}  & \textbf{56.8} & \underline{22.7} & \underline{54.4}  \\

\rowcolor{gray!20}
InternVL-2.0-SP-Pooling & InternViT‑300M  &
Qwen2‑0.5B &448 &  \textbf{58.4}  & \textbf{58.2} & \textbf{41.8} & \underline{85.0} & \textbf{1223.4} &  \underline{52.2} & \underline{56.6}  & \textbf{24.0} & \textbf{54.7} \\

\bottomrule

\end{tabular}%

\caption{\textbf{Methods Generalization.} We conducted experiments using the LLaVA-1.5 558k+665k training data. In the experiment of SP method applied to  InternVL-2.0, we only extract the visual spatial tokens from the original image.
}
\label{table:siglip}
\end{table*}


\begin{figure}[t]
  \centering
\includegraphics[width=1.0\linewidth]{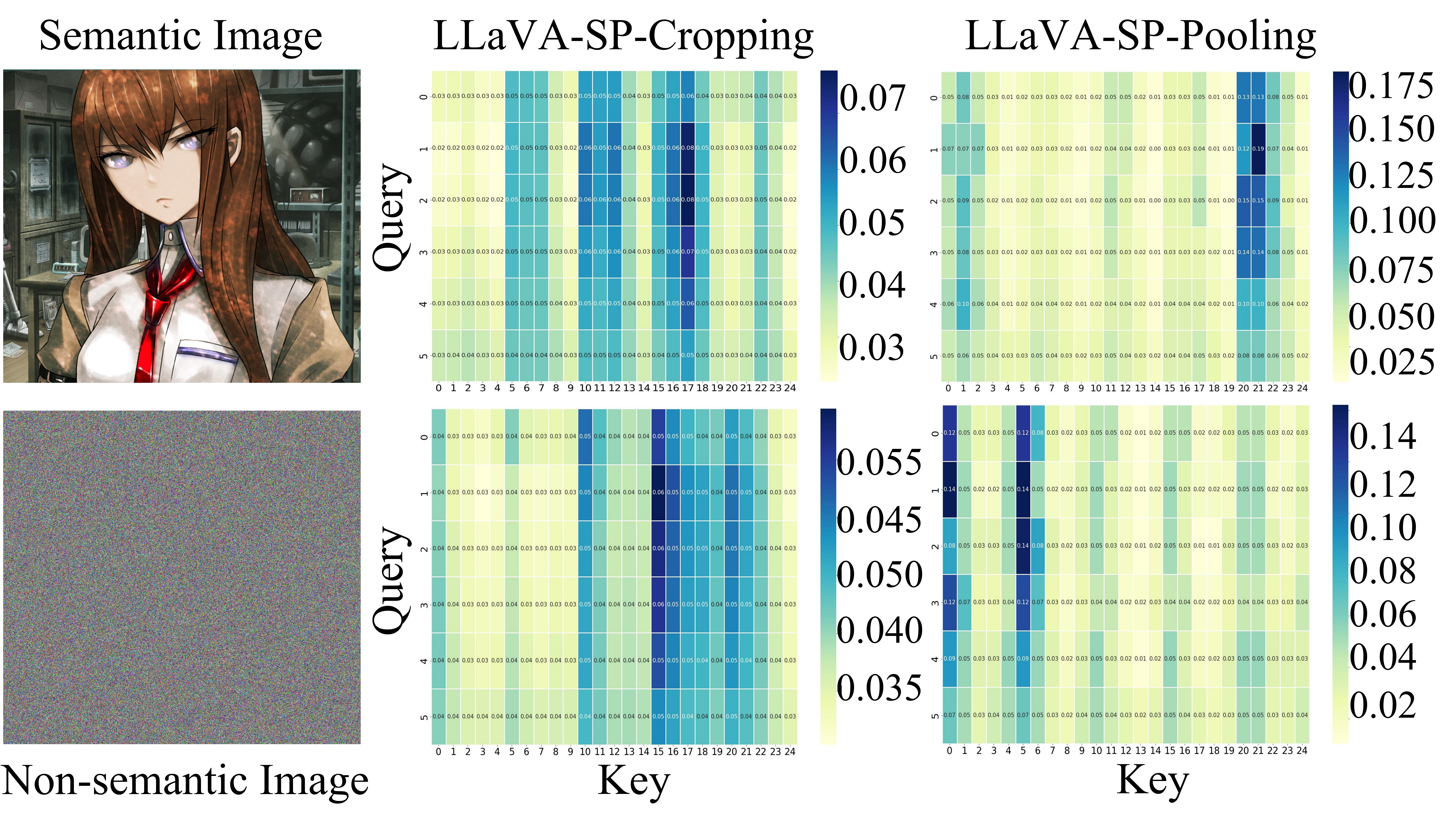}
   \caption{\textbf{Attention map visualization.} The vertical axis represents the queries, which consists of six visual spatial features $Z_{s\text{-}small}$, and the horizontal axis represents the keys, which is $Z_{s\text{-}big}$. The darker the color on the attention map, the higher the attention score. The attention score of LLaVA-SP-Cropping is more average, while LLaVA-SP-Pooling is more concentrated.}
   \label{attention}
\end{figure}

\noindent\textbf{Attention map visualization.} To validate the above hypothesis, we visualized the attention maps. As shown in \cref{attention}, the vertical axis of the attention map represents the queries, while the horizontal axis represents the keys. In LLaVA-SP-Cropping, attention is distributed more uniformly, whereas in LLaVA-SP-Pooling, it is more concentrated, focusing on only a few keys. For instance, in the Gaussian noise image at the bottom of \cref{attention}, where no significant regions exist, the attention score should be evenly distributed. However, the Pooling model excessively focuses on certain keys, with the highest attention score reaching 0.14, which is 7 times the minimum value, indicating an unreasonable distribution. This suggests that the pooling operation disrupts the original local features, hindering the model's ability to learn correct attention weights and leading it to overemphasize non-essential features, which harms visual feature fusion.


\begin{table}[t]

\footnotesize
\centering
\setlength{\tabcolsep}{3.85pt}
    \begin{tabular}{l|c|ccc|cc|c}
        \toprule
        Method & \multicolumn{1}{c|}{MME} & \multicolumn{3}{c|}{MMB} & \multicolumn{2}{c|}{SEED} & Avg\textsuperscript{N} \\
        & POS & SR & OL & PR & SR & IL  & \\
        \midrule

     LLaVA-1.5† &128.3  & 20.0 & 44.4 & 25.0 & 51.1 & 59.9 & 44.1 \\

        Honeybee~\cite{honeybee} & 116.7 & 15.6 & 42.0 & 54.2 & 43.5 & 54.4 & 44.7 \\
        DeCo~\cite{deco}& 116.7 & 24.4 & 48.1 & 41.7 & 46.6 & 58.5 & 46.3 \\

        LLaVA-SP-Pooling & \textbf{138.3} & 15.6 & 45.7 & 37.5 & \underline{49.0} & \underline{61.4} & \underline{46.4} \\

        \rowcolor{gray!20}
        LLaVA-SP-Cropping & 126.7 & \textbf{24.4} & \textbf{50.6} & 29.2 & \textbf{49.8} & \underline{61.7} & \textbf{46.5} \\
        \bottomrule
    \end{tabular}
    \caption{\textbf{Visual spatial understanding evaluation.} † indicates that the result is not reported in LLaVA-1.5~\cite{llava1.5}, and we tested the result using the official full-training parameter. 
    The abbreviations for task names denote Position (POS) in MME; Spatial Relationship (SR), Object Localization (OL) and Physical Relation (PR) in MMB; Spatial Relation (SR) and Instance Location (IL)in SEED-IMG. Our models fine-tuned with LoRA achieves the best score.}
    \label{tab:understand}
\end{table}

\subsection{Visual Understanding Enhanced Analysis}
\label{sec:vg}
\textbf{Visual spatial understanding.} We evaluated the model's visual spatial understanding capabilities, including precise visual localization,  fine-grained visual reasoning, and object relationship perception on MME~\cite{mme}, SEED-IMG~\cite{li2023seed} and MMBench~\cite{mmbench}. \cref{tab:understand} shows LLaVA-SP-Cropping achieves 46.5 in Avg\textsuperscript{N}, which is higher than DeCo~\cite{deco} and HoneyBee~\cite{honeybee}, both trained under the same configuration.

\begin{table}[t]

\footnotesize
\centering
\setlength{\tabcolsep}{1.5pt}
    \begin{tabular}{l|ccc| ccc| cc}
    \toprule
    Method & \multicolumn{3}{c|}{RefCOCO} & 
    
    \multicolumn{3}{c|}{RefCOCO+} & 
    
    \multicolumn{2}{c}{RefCOCOg} \\
     & val & test-A & test-B & val & test-A & test-B & val & test \\
    \midrule

    LLaVA-1.5†  & 54.7 &63.2  & 45.8 & 48.3 & 57.2 &37.8 & 50.8 & 50.6\\

    LLaVA-SP-Pooling & 60.0 & 69.3 & 47.7 & 55.2 & 65.4 & 42.8 & 55.2 & 56.4 \\

    \rowcolor{gray!20}
    LLaVA-SP-Cropping & \textbf{60.3}  & \textbf{69.7} & \textbf{47.8} & \textbf{55.4} & \underline{65.2} & \textbf{43.4}
    &\textbf{55.7} &\underline{56.1} \\
    \bottomrule
    \end{tabular}
\caption{\textbf{Visual grounding evaluation.}  † indicates that the results using the full-training LLaVA-1.5 official parameter. The experiments show that LLaVA-SP-Cropping performs best on fine-grained local image understanding tasks.}
    \label{tab:refcoco}
\end{table}

\noindent\textbf{Visual grounding.} The visual grounding task requires the model to output bounding boxes based on a given description. RefCOCO benchmark ~\cite{refcoco} evaluation results reflect the model's fine-grained local image understanding ability.  ~\cref{tab:refcoco} shows that our approaches greatly enhance visual grounding capability. LLaVA-SP-Cropping achieves the highest score, making it more suitable for tasks that require understanding fine-grained image details.

\noindent\textbf{Hallucination.} We evaluated the hallucination issue on POPE~\cite{pope} and MMVP~\cite{mmvp}. As shown in \cref{table:6}, both LLaVA-SP-Cropping and LLaVA-SP-Pooling achieve higher scores compared to LLaVA-1.5. Our methods effectively mitigate the CLIP-Blind problem~\cite{mmvp}, which refers to the inability of visual models to distinguish subtle differences between similar image pairs.


\begin{table}[t]
\small
\centering
\setlength{\tabcolsep}{4pt}
\begin{tabular}{l|cc}
\toprule
Method & MMVP & POPE \\
\midrule
LLaVA-1.5†       & 24.7 & 85.9 \\

\rowcolor{gray!20}
LLaVA-SP-Pooling   & \underline{30.7} & \textbf{86.5} \\
\rowcolor{gray!20}
LLaVA-SP-Cropping      & \textbf{31.3} & \underline{86.4} \\
\bottomrule
\end{tabular}
\caption{\textbf{Hallucination issue evaluation.} † indicates that the result using the official full-training parameter of LLaVA-1.5. Both LLaVA-SP-Cropping and LLaVA-SP-Pooling can alleviate the hallucination problem in MLLMs.}
\label{table:6}
\end{table}

\subsection{Methods Generalization} 

We replaced CLIP-ViT-L/14-336 with SigLIP-L/16-384 and applied SP method to InternVL-2.0. Our method focuses on enhancing the visual representation of CLIP, effectively adding an external module to CLIP. Other MLLMs, which involve higher resolutions, more visual tokens, and stronger vision encoder, are orthogonal to our approach, as their CLIP still has representational limitations. \cref{table:siglip} demonstrates that our approach can be adapted to stronger vision encoder and the novel MLLM framework.

\section{Conclusion}

In this work, we propose LLaVA-SP, which enhances the visual representation for MLLMs by adding only six visual spatial tokens to the original visual tokens. We propose a novel Projector, which uses convolutional kernels to extract visual spatial tokens and simulates two approaches for visual spatial ordering: ``from central region to global" and ``from abstract to specific". Additionally, we present two model variants to handle various visual understanding tasks. Finally, LLaVA-SP, fine-tuned with LoRA, outperforms other state-of-the-art methods on various benchmarks while maintaining nearly identical inference latency.

\section{Acknowledgements}
This work is supported in part by the National
Natural Science Foundation of China (Grant Nos.
62376034 and 92467105).

{
    \small
    \bibliographystyle{ieeenat_fullname}
    \bibliography{main}
}

\clearpage
\appendix
\section{Implementation details}
\label{sec:Implementation}

\noindent\textbf{Hyperparameters.} The experimental setup follows LLaVA-1.5~\cite{llava1.5}. The training strategy consists of pre-training and instruction tuning. In the pre-training stage, the LLM is frozen, and the projector is trained to align vision and language representation. In the instruction tuning stage, both the LLM and the projector are trained to enhance the model's ability to follow human instructions. Specific training hyperparameters are detailed in \cref{hyper}.

\begin{table}[h!]
    \centering
    \begin{tabular}{l|cc}
        \toprule
        Hyperparameter    & Pre-training                     & Instruction Tuning \\ \midrule
        Global batch size & 256                              & 128                \\
        Projector LR      & $1 \times 10^{-3}$               & $2 \times 10^{-5}$ \\
        LLM LR            &-                               & $2 \times 10^{-4}$ \\
        LR schedule       & \multicolumn{2}{c}{Cosine decay}                      \\
        Warmup ratio      & \multicolumn{2}{c}{0.03}                              \\
        LoRA rank         &-                               & 128                \\
        Optimizer         & \multicolumn{2}{c}{AdamW}                             \\
        Epoch             & \multicolumn{2}{c}{1}                                 \\
        Weight decay      & \multicolumn{2}{c}{0}                                 \\
        Deepspeed stage   & \multicolumn{2}{c}{2}                                 \\

        \bottomrule
    \end{tabular}
    \caption{\textbf{Training hyperparameters.} LR indicates learning rate.
    }
    \label{hyper}
\end{table}

\noindent\textbf{Details on Spatial Feature Extractor.}
We use convolutional kernels to extract six visual spatial tokens. The convolution input channels are 1024, which match the dimension of vision encoder outputs, and the output channels are 512. The total parameters of the projector are 1536 MiB. We also experimented with transformer blocks, a 4 layers encoder-decoder structure, where the input and output channels are both 1024. The total parameters of the projector are 836 MiB. As mentioned in the main paper, the convolutional kernels outperforms the transformer blocks.

\noindent\textbf{Details on Detail Feature Integrator.}
We use a simple linear layer with layer normalization to implement the Q and KV matrices for the cross-attention mechanism, where the input and output channel dimension are both 512.

\begin{figure}[htbp]
    \centering

    \begin{subfigure}{0.5\textwidth}
        \centering
        \includegraphics[width=\linewidth]{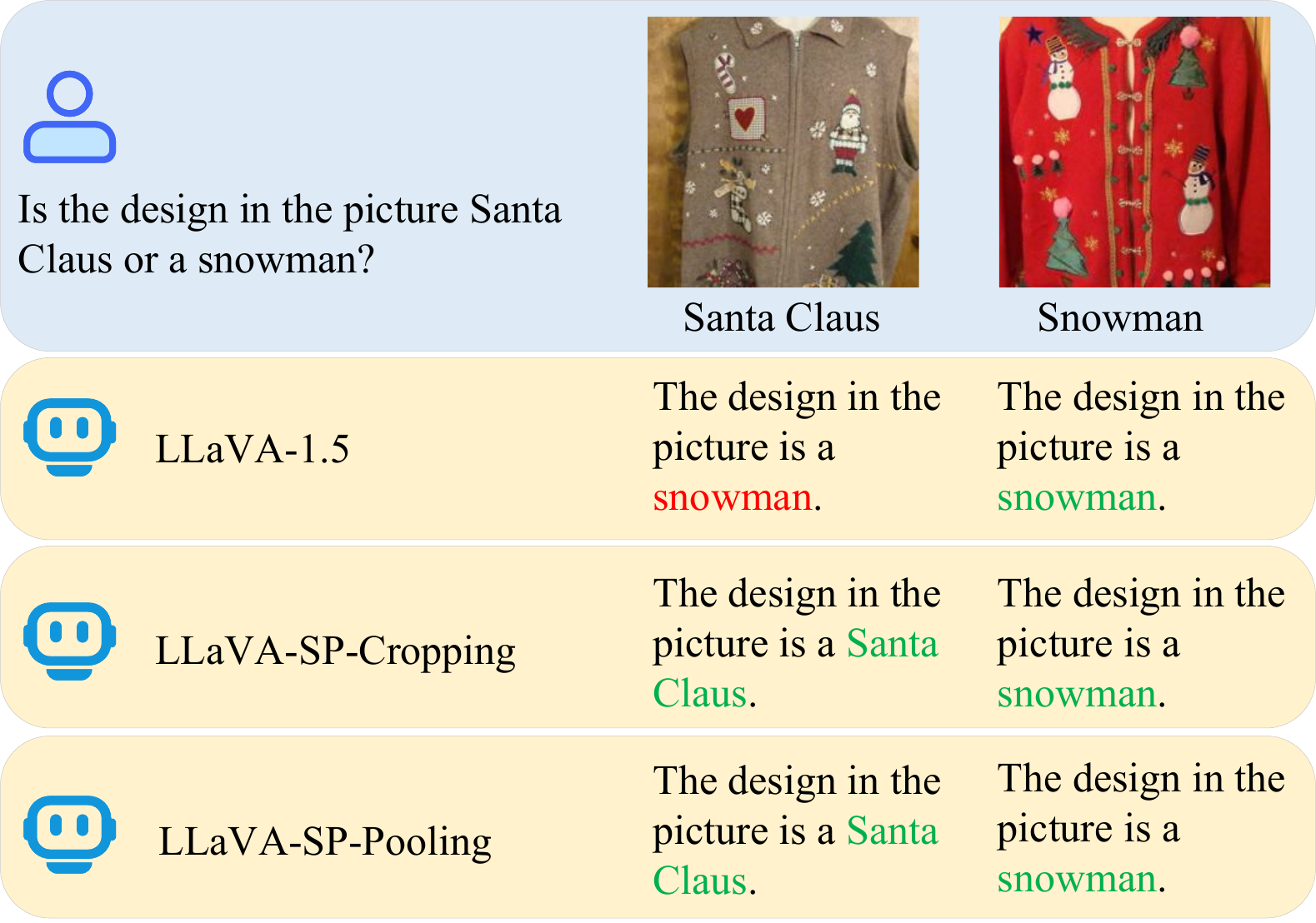} 
        \caption{Case 1.}
        \label{a}
    \end{subfigure}
    
    \vspace{0.5cm} 
    
    \begin{subfigure}{0.5\textwidth}
        \centering
        \includegraphics[width=\linewidth]{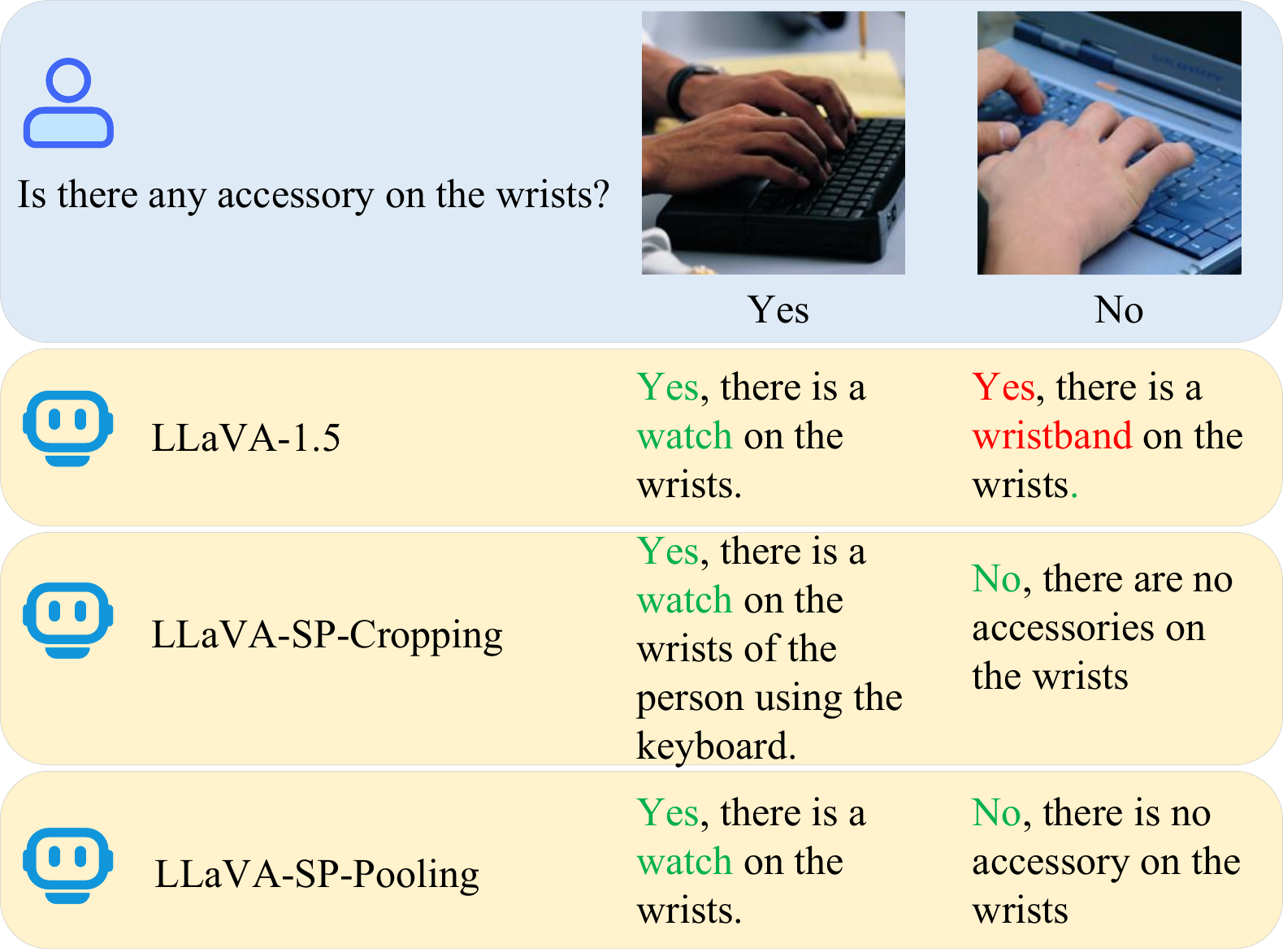} 
        \caption{Case 2.}
        \label{b}
    \end{subfigure}

    \caption{Examples of LLaVA-SP-Cropping and LLaVA-SP-Pooling on MMVP dataset. {\color{ForestGreen} Correct} and {\color{red} incorrect} answers are marked in {\color{ForestGreen} green} and {\color{red} red} respectively.}
    \label{fig.mmvp}
\end{figure}

\section{Qualitative Results}

\subsection{Case Study on MMVP Benchmark}
We provide a case study of LLaVA-SP (comprising LLaVA-SP-Cropping and LLaVA-SP-Pooling) on the MMVP~\cite{mmvp}, to investigate their enhanced capabilities compared to the base LLaVA-1.5. From the output answers, we observe that: 1) In \cref{a}, LLaVA-1.5 incorrectly identified both designs as a snowman. For example, when presented with a sweater design featuring Santa Claus-like elements such as a red color scheme, white trim, and Santa-like patterns, LLaVA-1.5 still answered that it was a snowman. In contrast, both LLaVA-SP-Cropping and LLaVA-SP-Pooling correctly identified the design as Santa Claus. The SFE in LLaVA-SP-Cropping likely focused on detailed regional features. For instance, it could have zeroed in on the specific Santa-like patterns and the red-white color combination that is characteristic of Santa Claus designs. Similarly, LLaVA-SP-Pooling, with its pooling operation in SFE, captured the overall visual context effectively. It could recognize the combination of elements that are typical of Santa Claus designs, rather than misinterpreting them as those of a snowman.
2) In \cref{b}, when determining whether there are accessories on wrists, LLaVA-1.5 made mistakes. For example, in an image where a person was using a keyboard and had a watch on their wrist, LLaVA-1.5 incorrectly stated that there were no accessories on the wrist. Both LLaVA-SP-Cropping and LLaVA-SP-Pooling demonstrated superiority. LLaVA-SP-Cropping was able to extract relevant visual cues through its cropping-based SFE. It could focus on the wrist area and accurately identify the presence of the watch. LLaVA-SP-Pooling also performed well. Its pooling-based SFE captured the overall visual context, and DFI helped in integrating relevant features. This enabled it to correctly identify the presence of accessories on the wrist.

Overall, through the innovative designs of SFE and DFI, LLaVA-SP can distinguish differences between ``CLIP-bind pairs" images that CLIP perceives as similar despite their clear visual differences.

\begin{figure*}[htbp]
    \centering

    \begin{subfigure}{0.9\textwidth}
        \centering
        \includegraphics[width=\linewidth]{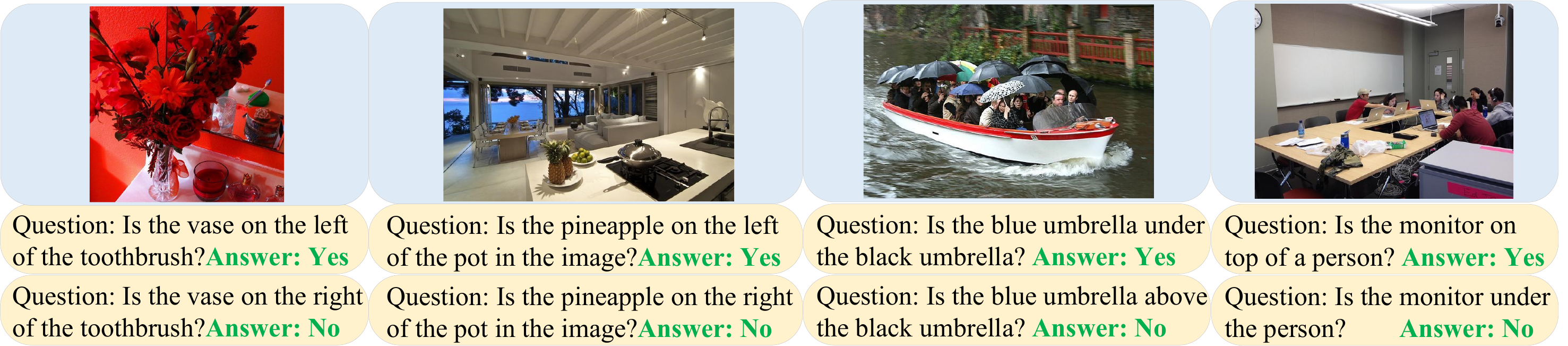} 
        \caption{Position recognition task.}
        \label{pos}
    \end{subfigure}

    \vspace{0.5cm} 

    \begin{subfigure}{0.9\textwidth}
        \centering
        \includegraphics[width=\linewidth]{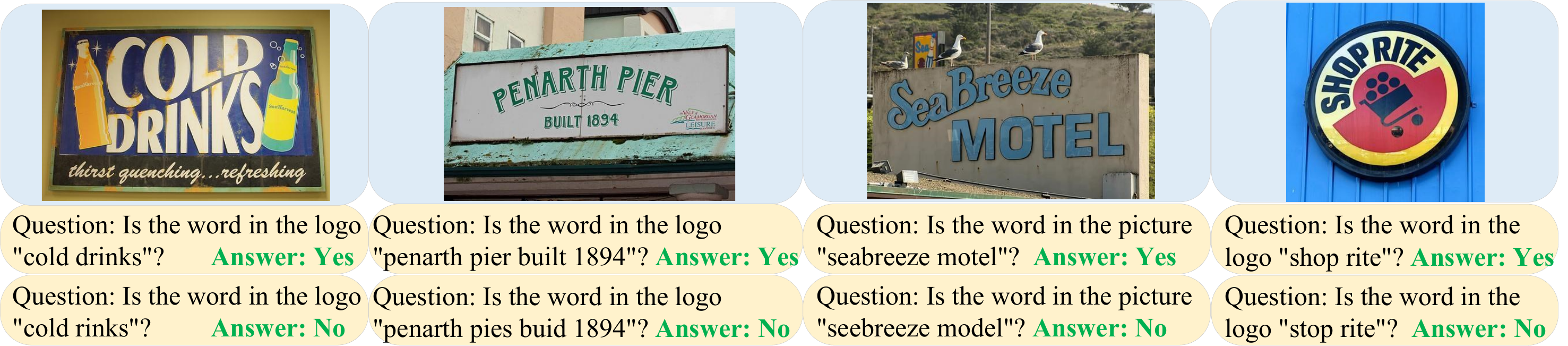} 
        \caption{OCR task.}
        \label{ocr}
    \end{subfigure}

    \vspace{0.5cm} 

    \begin{subfigure}{0.9\textwidth}
        \centering
        \includegraphics[width=\linewidth]{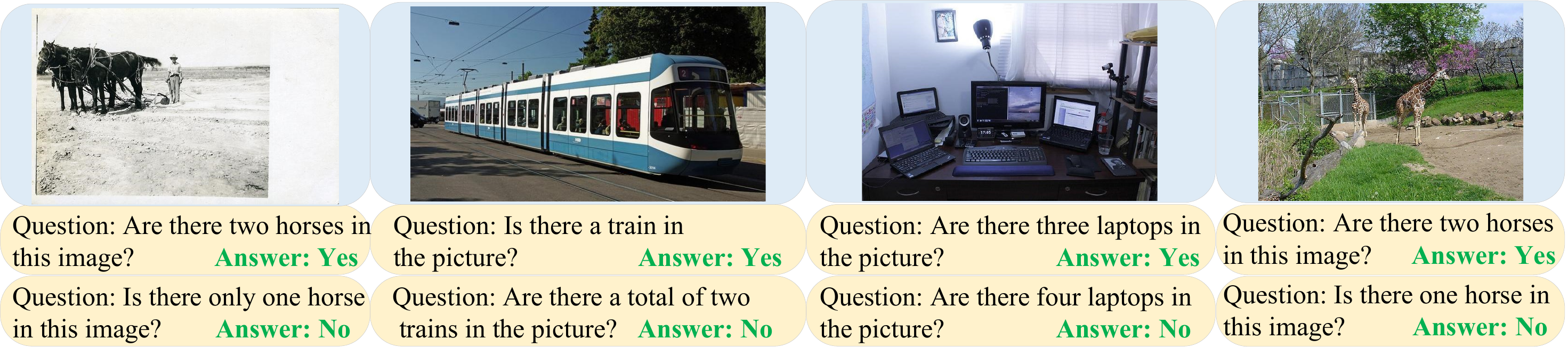} 
        \caption{Counting task.}
        \label{count}
    \end{subfigure}

    \vspace{0.5cm} 
    \begin{subfigure}{0.9\textwidth}
        \centering
        \includegraphics[width=\linewidth]{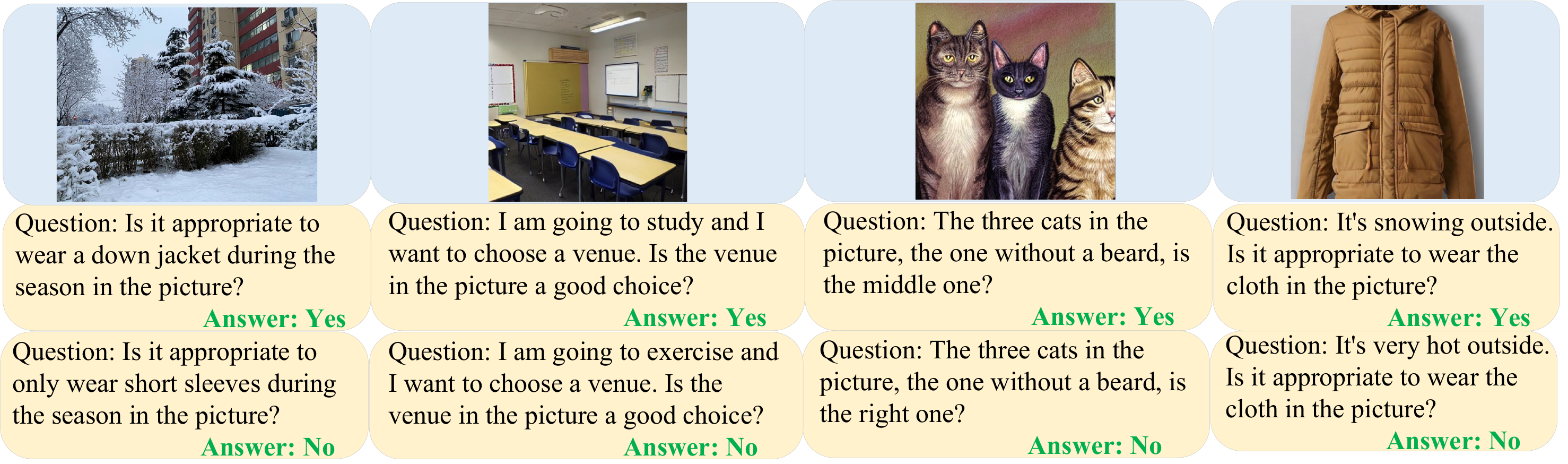} 
        \caption{Commonsense reasoning task.}
        \label{reason}
    \end{subfigure}
    \caption{Examples of LLaVA-SP-Cropping and LLaVA-SP-Pooling on MME dataset.}
    \label{fig:mme}
\end{figure*}

\subsection{Case Study on MME Benchmark}

We provide a case study of LLaVA-SP (comprising LLaVA-SP-Cropping and LLaVA-SP-Pooling) on the MME~\cite{mme}. From the output answers, we observe that: 1) In the position recognition task, LLaVA-SP demonstrated excellent performance. It was able to accurately determine the spatial relationships between objects. For example, in \cref{pos}, faced the question ``Is the pineapple on the left of the pot in the image?", LLaVA-SP can accurately analyze the visual layout. When asked another question about the same image, ``Is the pineapple on the right of the pot in the image", LLaVA-SP again precisely assesses the position. The SFE dissects the visual scene, focusing on the relative positions of the pineapple and the pot. This ability to handle multiple position-related questions about a single image accurately and consistently is a testament to the superiority of LLaVA-SP.
2) In the OCR task, LLaVA-SP effectively recognized text in images, accurately identifying words in logos. For example, in \cref{ocr}, when presented with an image of a drinks shop logo that has the text ``COLD DRINKS" in a unique cursive font, LLaVA-SP's SFE first pinpoints the text region by detecting the contrast between the lighter text and the darker background. It then zooms in on each letter. For the letter ``C", it carefully analyzes the curved shape, the smooth transition of the stroke, and the way it connects to the following letter ``O". The SFE is able to handle the complexity of the cursive style and the decorative elements. The DFI further refines the recognition. It picks up on the minute variations in the thickness of the strokes and the small loops in the letters. This enables LLaVA-SP to accurately recognize ``COLD DRINKS". 
3) In counting tasks, LLaVA-SP provided accurate counts. For example, in \cref{count}(c) when presented with the image and the question ``Are there three laptops in the picture?" LLaVA-SP's SFE scanned the image, identifying the laptops based on their characteristic shapes and visual patterns. It differentiated laptops from other objects. The DFI then enhanced the analysis by focusing on details like the screen bezels and keyboard markings. This allowed LLaVA-SP to precisely count the laptops, answering both ``Are there three laptops in the picture?" and ``Are there four laptops in the picture?" accurately. It could handle occlusions and variations in laptop appearances, outperforming models that might miscount or miss some laptops.
4) In commonsense reasoning tasks, LLaVA-SP exhibited strong reasoning capabilities, offering correct answers to situational questions. For example, in \cref{reason}, when presented with an image and asked ``It's snowing outside. Is it appropriate to wear the cloth in the picture?" and ``It's very hot outside. Is it appropriate to wear the cloth in the picture?", LLaVA-SP's SFE analyzes the visual details of the cloth. It focuses on characteristics such as thickness and material texture. In snowy conditions, the SFE recognizes that the cloth appears warm and suitable. The DFI then further refines these features by integrating fine-grained details like the style and any associated accessories that suggest cold-weather wear. For the hot weather question, LLaVA-SP realizes the cloth is too heavy and inappropriate.


\section{Qualitative Analysis of LLaVA-SP}

To evaluate the effectiveness of LLaVA-SP in visual understanding, we qualitatively analyze its performance in comparison with LLaVA. The analysis highlights the strengths of SFE's design, including cropping-based strategy, which enhances the model's ability to capture fine-grained regional details, and pooling-based strategy, which enables adaptive global reasoning. Two example images (Figure~\ref{sleep} and Figure~\ref{boat}) are used for this analysis.

In Figure~\ref{sleep}, a desk scene with objects such as a laptop, books, and a chair is presented. LLaVA produces a general description, but it makes errors in object recognition, such as misidentifying the number of potted plants and failing to recognize the specific spatial arrangement of items. In contrast, LLaVA-SP generates more accurate and detailed descriptions. Specifically, using SFE with a cropping-based strategy, LLaVA-SP correctly identifies the presence and positions of key objects like the laptop and books. Meanwhile, with the pooling-based strategy, it focuses on summarizing the scene's overall calm and organized atmosphere, emphasizing its ability to balance detail with context.

Similarly, in Figure~\ref{boat}, a beach scene is depicted, featuring people, boats, and trees. LLaVA fails to detect small objects like the boats and provides an inaccurate count of individuals in the scene, estimating at least 11 people when there are fewer. On the other hand, LLaVA-SP demonstrates significant improvements. The cropping-based strategy identifies the small boats and describes interactions between people and nearby objects with high precision. In contrast, the pooling-based strategy captures the overarching aesthetic of the beach scene, such as the presence of boats in the water and the relaxing coastal environment.

The superior performance of LLaVA-SP can be attributed to the combined effects of SFE and DFI. The cropping-based approach leverages SFE to focus on localized regions, making it effective for tasks that require object-level recognition. In contrast, the pooling-based approach benefits from DFI to aggregate information globally, providing a more abstract understanding of the scene. These two strategies offer complementary strengths, allowing LLaVA-SP to excel in both fine-grained and holistic reasoning.

In summary, LLaVA-SP significantly outperforms LLaVA in capturing both regional details and global context. The cropping-based approach is particularly suitable for precise object-level analysis, while the pooling-based approach excels in summarizing scene-level information. Together, they demonstrate the versatility and robustness of LLaVA-SP in handling diverse visual reasoning tasks.

\section{Deep Analysis between LLaVA-SP-Cropping and LLaVA-SP-Pooling}

\begin{figure*}[htbp]
    \centering

    \begin{subfigure}{0.9\textwidth}
        \centering
        \includegraphics[width=\linewidth]{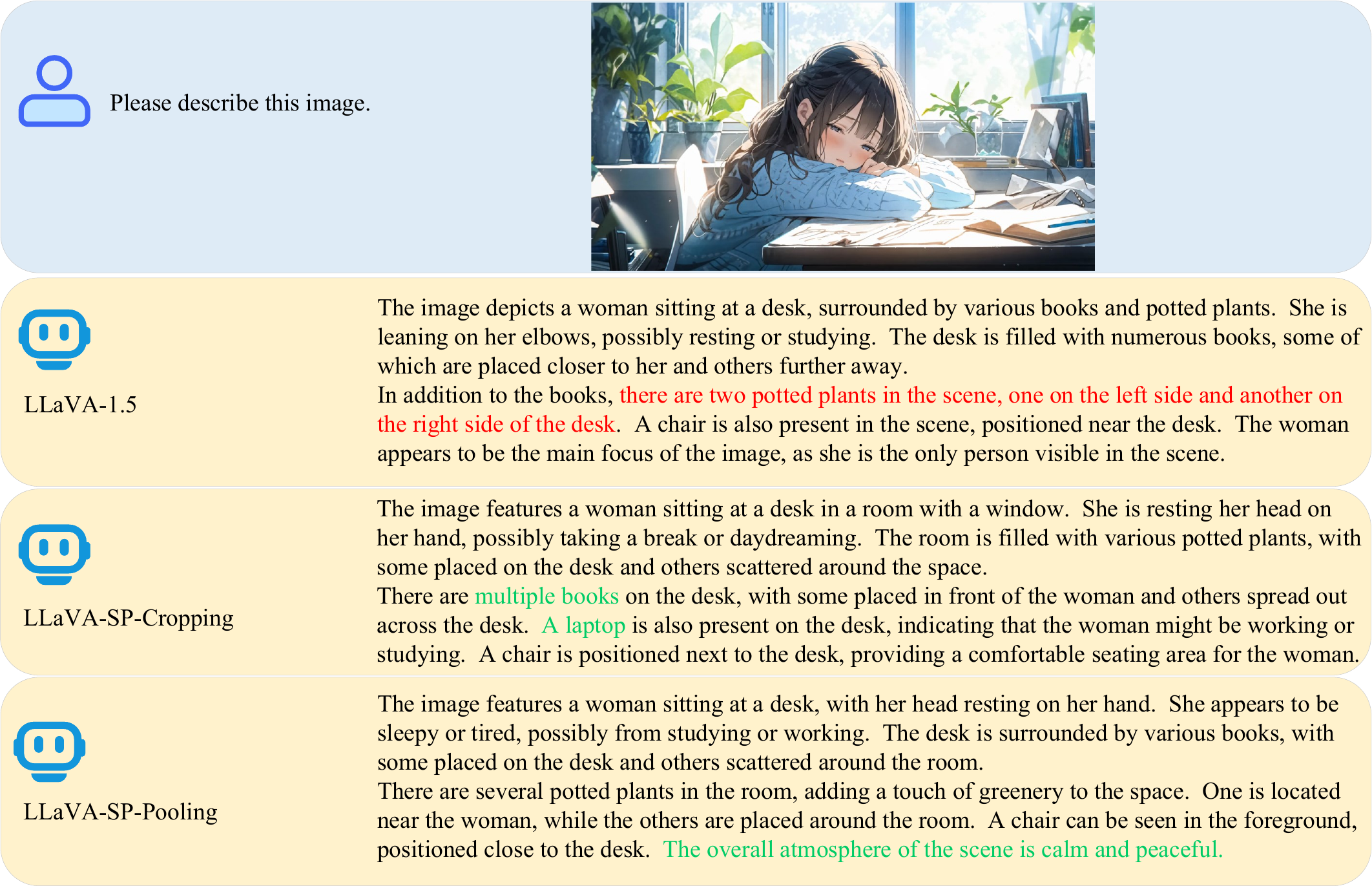} 
        \caption{LLaVA-1.5 mistakenly pointed out that there are only two plants in the image. LLaVA-SP-Cropping identified details in the image, such as the books and laptop. LLaVA-SP-Pooling captured the overall atmosphere of this image.}
        \label{sleep}
    \end{subfigure}

    \vspace{0.5cm} 
    
        \begin{subfigure}{0.9\textwidth}
        \centering
        \includegraphics[width=\linewidth]{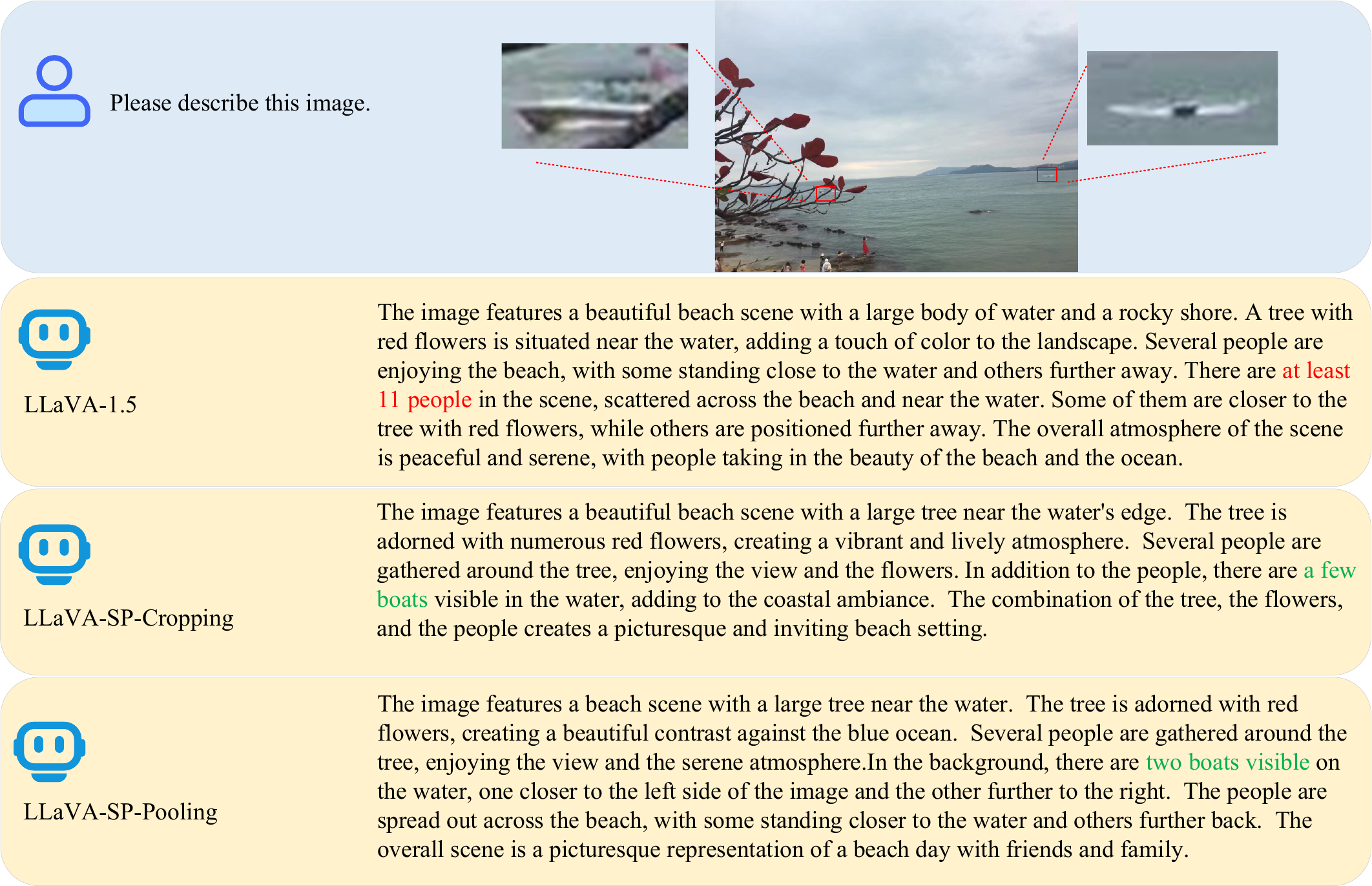} 
        \caption{LLaVA-SP-Cropping and LLaVA-SP-Pooling all detected the small boat in the image, whereas LLaVA-1.5 failed to describe the small boat and incorrectly stated the number of people.}
        \label{boat}
    
    \end{subfigure}

    \caption{Deep Analysis between LLaVA-SP-Cropping, LLaVA-SP-Pooling and LLaVA-1.5. 
 {\color{ForestGreen} Correct} and {\color{red} incorrect} answers are marked in {\color{ForestGreen} green} and {\color{red} red} respectively.}
    
    \label{3}
\end{figure*}

\begin{figure*}[htbp]
    \centering
    \includegraphics[width=0.9\linewidth]{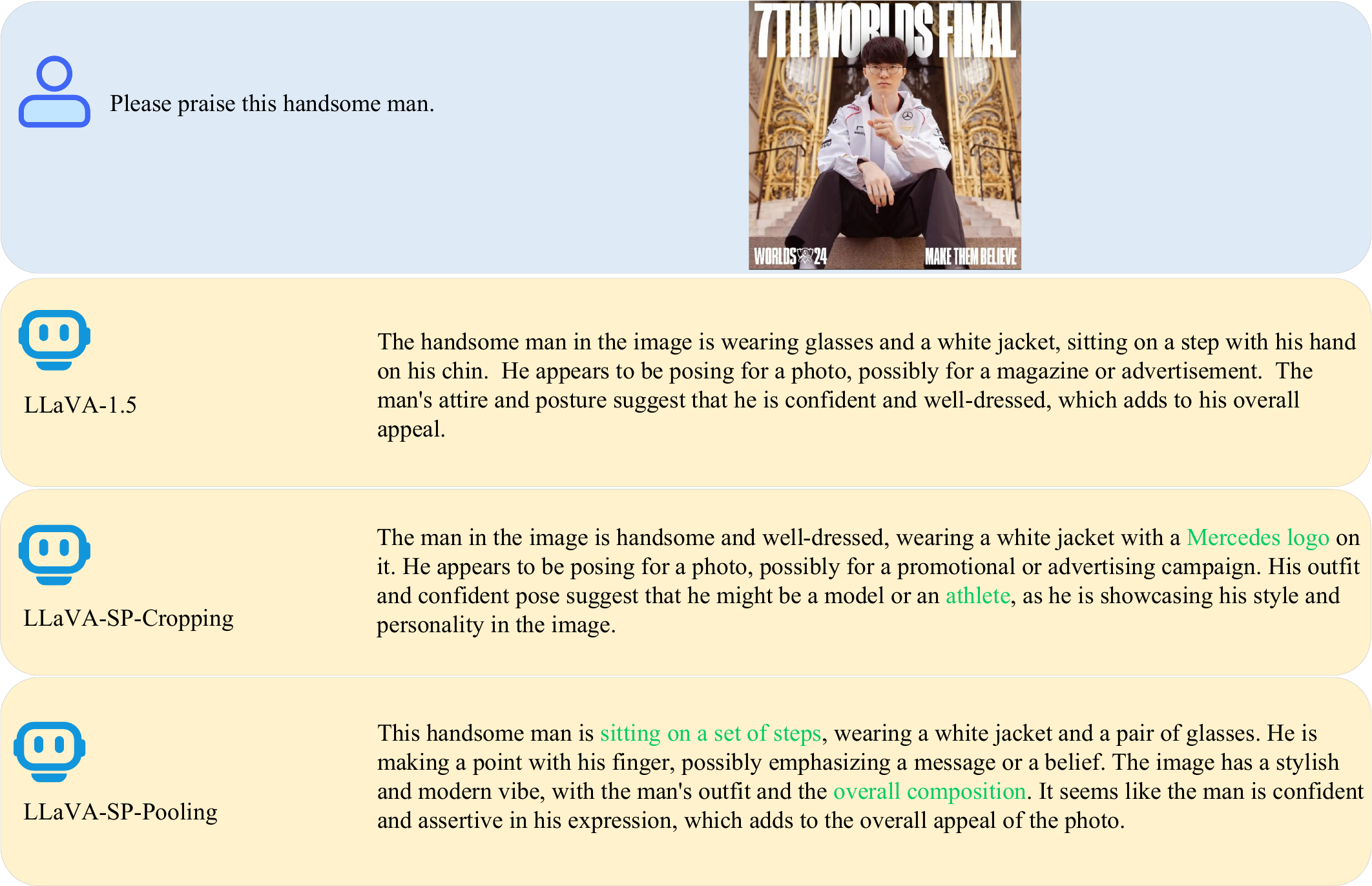}
    \caption{Compared to LLaVA-1.5, LLaVA-SP-Cropping captured the Mercedes logo on Faker's clothing, while LLaVA-SP-Pooling focused on the overall composition.}
    \label{4}
\end{figure*}

The SFE enhances the vision encoder by introducing six visual spatial tokens, which can be obtained through two distinct methods: cropping and pooling. While both approaches aim to enrich the visual representation, their focus and mechanisms differ significantly, leading to distinct performance advantages in different scenarios. In this section, we conduct a deep analysis of LLaVA-SP-Cropping and LLaVA-SP-Pooling, comparing their outputs across the examples shown in ~\cref{3,4}. 

\noindent\textbf{Cropping Method:} The cropping approach extracts regional features by progressively narrowing the focus of the ViT patch features, starting from the global context and cropping inward toward the central region. This process generates multi-scale features arranged in the order of `central region to global,’ ensuring that detailed information from small but crucial regions is prioritized. For example, in ~\cref{sleep}, LLaVA-SP-Cropping accurately captures specific details such as the posture of the individual and the texture of the surrounding elements, which are critical for precise understanding. Similarly, in ~\cref{boat}, cropping captures fine-grained features like the intricate design of objects and their interactions with the environment, showcasing its effectiveness in reasoning at a detailed level. Lastly, in ~\cref{4}, cropping excels at identifying the specific Mercedes logo on the clothing, a region-specific detail that might otherwise be overlooked in a global representation.

\noindent\textbf{Pooling Method:} In contrast, the pooling approach adopts a hierarchical strategy, using adaptive pooling layers to generate multi-scale features that range from abstract to specific. These features are arranged sequentially, first capturing the global structure and then transitioning to finer details, mimicking the way humans perceive visual scenes~\cite{autogressiveimage}. Pooling is particularly effective in scenarios requiring a holistic understanding. For instance, in ~\cref{sleep}, pooling captures the overall context of the scene, emphasizing the arrangement and spatial composition of the surroundings. In ~\cref{boat}, pooling highlights the broader interactions between objects, such as the relationship between the primary elements and the background environment. Finally, in ~\cref{4}, pooling focuses on the subject's confident posture and overall compositional balance, offering a cohesive interpretation of the entire image.

\noindent\textbf{Comparative Analysis:} The differences between cropping and pooling stem from their respective focuses. Cropping excels in tasks requiring fine-grained image understanding, as it isolates and highlights specific regions with high precision. This is evident in examples such as the distinct feature in ~\cref{4} and the detailed object interactions in ~\cref{boat}. Pooling, on the other hand, is better suited for tasks demanding a comprehensive understanding of the scene, as it captures the global structure and integrates contextual information. This is particularly beneficial in scenarios like ~\cref{sleep}, where pooling effectively conveys the overall spatial arrangement and mood of the scene.

Both methods leverage the same foundational SFE mechanism, reshaping ViT patch features into their original 2D structures before processing them into multi-scale features. However, their strategies for organizing these features (``from central region to global" for cropping and ``from abstract to specific" for pooling) lead to distinct strengths. Cropping prioritizes regional detail, making it more effective in capturing small but critical features. Pooling, by focusing on abstract-to-specific hierarchical information, provides a more holistic understanding of the image. Together, these methods complement each other, offering a flexible framework for addressing both fine-grained and global visual reasoning tasks.

\section{Limitation and Future Work}
1) Did not utilize larger-scale LLMs:
The experiments were conducted only on LLMs with 7B parameters, and the effectiveness of the method has not been validated on larger-scale LLMs. Future work will involve experiments on various LLMs such as Qwen2.5, Mistral, and LLaMA3.

\noindent2) Large Parameters:
The SFE employs convolutional kernels to extract spatial information from visual features. The input and output channels  of the convolutional kernels are 1024 (equal to the visual feature dimension output by ViT) and 512, respectively. Additionally, large kernels with a size of 16 are used. These will lead to a large number of parameters. According to the convolutional parameter calculation formula, the parameters for each convolutional kernel is:
\begin{equation}
parameters = C_{in}\times C_{out}\times Width\times
Height,
\label{eq:attention}
\end{equation}

\noindent where the $C_{in}$ and $C_{out}$ denotes the input channels and output channels respectively, $Width$ and $Height$ denotes the size of convolutional kernel.

In the future, we will adopt more efficient model design approaches to reduce parameters, improve training and inference speed, and achieve a trade-off between model performance and efficiency.

\clearpage

\end{document}